\documentclass[letterpaper]{article}
\usepackage{aaai}
\usepackage{times}
\usepackage{hyperref}
\usepackage{hyperref}
\hypersetup{hidelinks,
	colorlinks=true,
	allcolors=black,
	pdfstartview=Fit,
	breaklinks=true}
\usepackage{helvet}
\usepackage{courier}
\usepackage{multirow}
\usepackage{colortbl}
\usepackage[table]{xcolor}
\usepackage{booktabs}
\usepackage{amsmath}
\usepackage{amssymb}
\usepackage{pifont}
\usepackage{tabularray}
\usepackage{graphicx} 
\usepackage{natbib}  
\usepackage{caption} 
\DeclareCaptionStyle{ruled}{labelfont=normalfont,labelsep=colon,strut=off} 
\usepackage{algorithm}
\usepackage{algorithmic}
\usepackage{bibentry}

\begin{document}
%
\title{DHGS: Decoupled Hybrid Gaussian Splatting for Driving Scene}
\author{Xi Shi$^1$\thanks{These authors contributed equally.}, Lingli Chen$^{1\ast}$, Peng Wei$^{1,\,2*}$, Xi Wu$^{1}$\thanks{Corresponding author.}, Tian Jiang$^{1}$,
Yonggang Luo$^{1,\,2}$, Lecheng Xie$^{1}$\\
$^{1}$Changan Auto, AILab\\
$^{2}$State Key Laboratory of Intelligent Vehicle Safety Technology
\\
\{shixi, chenll4, weipeng4, wuxi, jiangtian, luoyg3, xielc\}@changan.com.cn\\
}

\maketitle
\begin{abstract}
\begin{quote}
Existing Gaussian splatting methods often fall short in achieving satisfactory novel view synthesis in driving scenes, primarily due to the absence of crafty designs and geometric constraints for the involved elements. This paper introduces a novel neural rendering method termed Decoupled Hybrid Gaussian Splatting (DHGS), targeting at promoting the rendering quality of novel view synthesis for static driving scenes. The novelty of this work lies in the decoupled and hybrid pixel-level blender for road and non-road layers, without the conventional unified differentiable rendering logic for the entire scene. Still, consistency and continuity in superimposition are preserved through the proposed depth-ordered hybrid rendering strategy. Additionally, an implicit road representation comprised of a Signed Distance Function (SDF) is trained to supervise the road surface with subtle geometric attributes. Accompanied by the use of auxiliary transmittance loss and consistency loss, novel images with imperceptible boundary and elevated fidelity are ultimately obtained. Substantial experiments on the Waymo dataset prove that DHGS outperforms the state-of-the-art methods. The project page where more video evidences are given is:  \href{https://ironbrotherstyle.github.io/dhgs\_web}{https://ironbrotherstyle.github.io/dhgs\_web}.
\end{quote}
\end{abstract}

\begin{figure}[t]
\centering
\includegraphics[width=0.9\columnwidth]{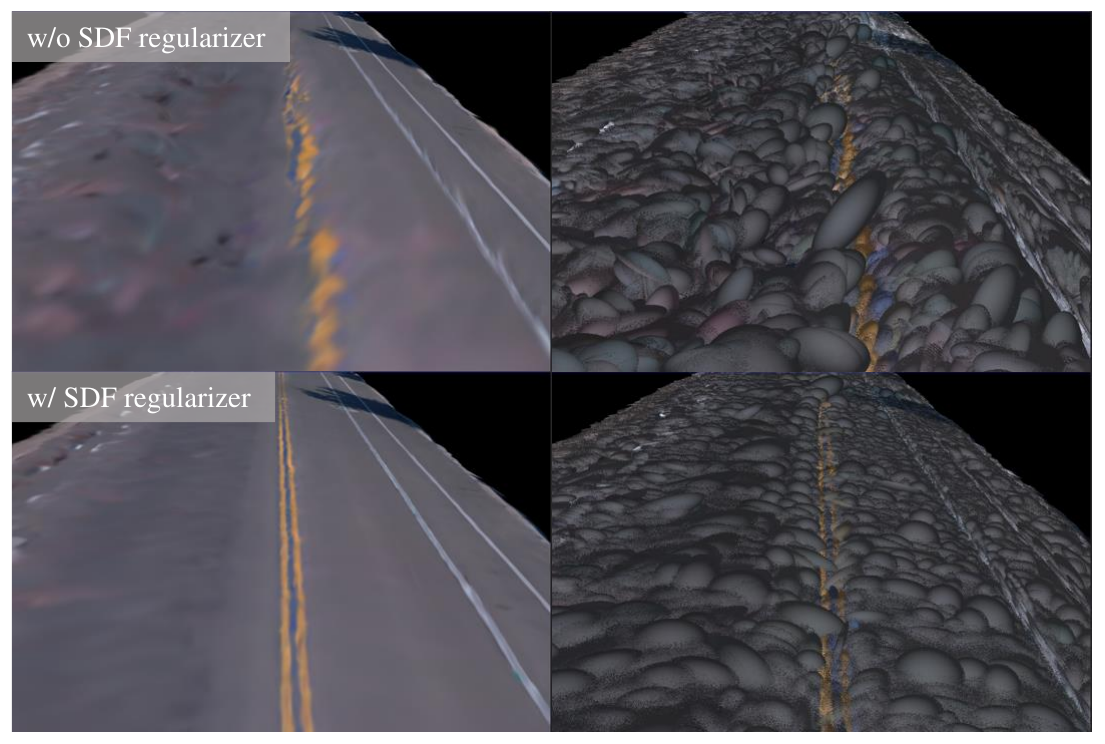} 
\caption{The comparison of the rendered road images and ellipsoids on novel view synthesis in Waymo dataset, with the top row displaying results without SDF regularizer, and the bottom row showcasing results obtained from the road model using SDF regularizer. It can be observed that the inclusion of SDF regularization leads the road model to render higher-quality images with the help of enhanced road geometry under decoupled scene representation.}
\vspace{-15pt}
\label{fig:ellp-comparison}
\end{figure}

\section{Introduction}
Autonomous Driving (AD) is a data-driven task that has attracted the attention of the 3D reconstruction community for many years. Due to the expense of collecting a tremendous amount of data, the neural rendering technique which mainly consists of Neural Radiance Fields (NeRFs) \cite{mildenhall2020nerf, barron2021mip} and Gaussian Splatting \cite{kerbl20233d} emerges as an alternative.
With the gathered AD data, a synthetic driving scenario can be effectively generated, thereby facilitating closed-loop simulation and testing, with a variety of sensor configurations that are not typically achievable in real-world tests \cite{yang2023unisim, tonderski2024neurad, lindstrom2024nerfs}. Compared with NeRF-based approaches, researchers utilizing Gaussian splatting can train and deploy a model more efficiently and flexibly. Despite the significant advancements in scene reconstruction offered by Gaussian splatting, there are still imperfections in the novel view synthesis capabilities, especially for critical downstream perceptual tasks such as online mapping \cite{li2023lanesegnet} of AD scenes. Current methods either model the whole driving scene in a consolidated manner \cite{ kerbl20233d, huang20242d, cheng2024gaussianpro, lu2024scaffold}, necessitating a uniform growth and pruning strategy for all scene elements, i.e. road, building, and distant views, or model the near scene and far sky \cite{ wu2023mars, EDUS} dividually. These approaches tend to emphasize the overall or specific distant elements, often overlooking the nearby synthesis quality which is brittle and remarkably impacted by the change of camera perspective. 
We believe that prioritizing the road is essential, as its geometric properties are fundamental to the success of AD emulation systems.

In this paper, we propose a static neural rendering method named Decoupled Hybrid Gaussian Splatting for Driving Scene (DHGS) to enhance the rendering quality of novel view synthesis, particularly for nearby surroundings with fewer artifacts and more subtle details. The insight of the method comes from the existing geometric prior that could be utilized to supervise the nearby road elements along with other non-road environments in an innovative pattern. Specifically, we decouple the whole driving scene into a nearby road and an environment part represented by two distinct Gaussian models, considering optimizing the two models based on their geometric characters accordingly. Before that, point clouds are projected and segmented using semantic 2D masks to derive the initial road and environment point clouds. The road point cloud is exclusively used for pre-training the implicit Signed Distance Function (SDF), which serves as the supervision of road training.
The overall training process is conducted via the introduced depth-ordered hybrid rendering strategy by the blender of rendered images of both models, assisted with transmittance loss that regularizes the accumulated transmittance of each componential Gaussian property, as well as consistency loss which bonds the two areas compactly. Besides, we refine the road elements using the SDF constraint that is expressly designed for the surface regularization, achieving uniform and orderly road Gaussians. As depicted in Figure \ref{fig:ellp-comparison}, a robust geometric prior enables the reconstructed scene to retain high-quality structural information even under significant perspective transformations. This capability is of paramount importance for the simulation system to generate hyper-realistic and authentic visual input, which promotes the vehicle to make well-informed decisions in complex driving scenarios.
To summarize, the contributions of the proposed DHGS are listed as follows:

\begin{itemize}
\item[$\bullet$]We first propose to decouple the modeling of driving scene into near road model and other non-road
model, enabling the separate optimization of two models and rendering jointly with the designed depth-ordered hybrid rendering strategy.
 
\item[$\bullet$]We propose an implicit road representation for enhanced guidance of road Gaussians, complemented by transmittance and consistency losses ensuring continuity and consistency.

\item[$\bullet$] Extensive quantitative and qualitative experiments on the popular Waymo dataset demonstrate that DHGS achieves the state-of-the-art performance on scene reconstruction and diversified novel free-view synthesis.
\end{itemize}

\section{Related Work}
\subsection{Neural Radiance Field Representations}
NeRF \cite{mildenhall2020nerf} has garnered remarkable attention within the neural rendering domain, pioneering the use of a Multi-Layer Perceptron (MLP) to model 3D scenes implicitly. The computationally intensive neural volume rendering and the finite number of sampling points hinder the deployment of NeRF in intricate and expansive driving scenarios. 
 Mip-NeRF \cite{barron2021mip} proposes a more efficient sampling strategy, refining the sampling region from a single ray to a frustum of a viewing cone. Both Mip-NeRF 360 \cite{barron2022mip} and NeRF++ \cite{zhang2020nerf++} segment the scene into proximal and distal zones, employing warping operations on the latter to handle the unbounded 3D scene. 
 Instant-NGP \cite{muller2022instant} leverages a hash grid for feature storage and streamlines the MLP size to release the computational burden. Plenoxels \cite{fridovich2022plenoxels} employs sparse grids to reduce storage space and utilizes Spherical Harmonics (SH) to represent the appearance. These approaches rely on data from 360-degree object-centric trajectories, while free and long driving scenes draw less attention. F2-NeRF \cite{wang2023f2} proposes a novel space-warping method that supports arbitrary input camera trajectories, enabling the reconstruction of driving scenes. StreetSurf \cite{guo2023streetsurf} extends prior object-centric neural surface reconstruction techniques to address the unique challenges inherent in unbounded street perspectives, demonstrating promise for a range of subsequent downstream tasks.  
Despite the notable advancements in ray marching and the innovative designs of MLPs, 
 these methods remain a trade-off between rendering fidelity and efficiency, which may restrict their utility in expansive autonomous driving scenarios.

\subsection{Gaussian Splatting Representations}
3DGS \cite{kerbl20233d} is a seminal work that innovates the use of 3D Gaussian ellipsoids to explicitly represent a scene and leverages CUDA for parallel rendering, thus surpassing NeRFs in both rendering quality and efficiency.
HUGS \cite{zhou2024hugs} enables real-time rendering of new perspectives, delivering both 2D and 3D semantic information with remarkable precision.
\begin{figure*}[t]
\centering
\includegraphics[width=1.95\columnwidth]{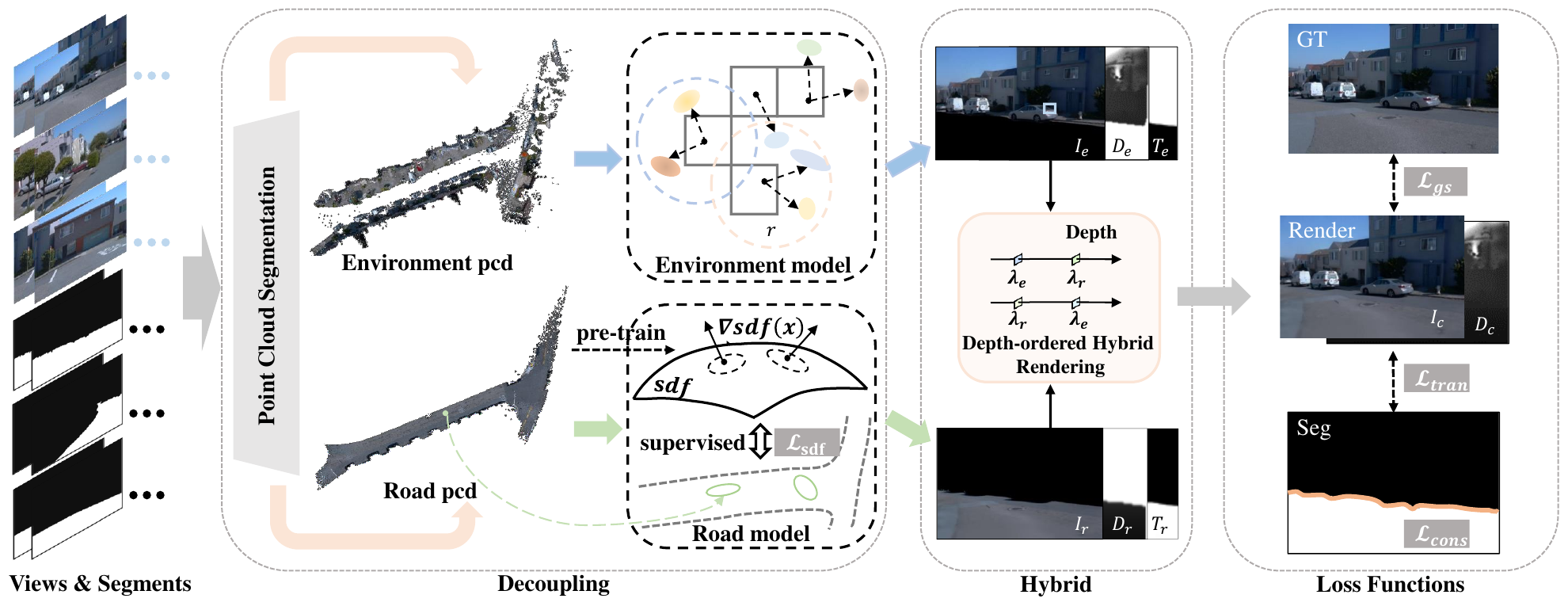} 
\caption{The training pipeline of the proposed method for driving scene reconstruction. Given consecutive multi-camera images along with their respective road and non-road masks, we initially generate decoupled road pcd (point cloud) and environment pcd, a road SDF is then pre-trained as subsequent guidance for the road Gaussian model. The environment pcd enables the initialization for the environment Gaussian model producing $I_e$, which is composed to the rendered image $I_c$ with image $I_r$ from paratactic road model via the proposed depth-ordered hybrid rendering.}
\label{fig:pipeline}
\vspace{-10pt}
\end{figure*}
The vanilla 3DGS focuses solely on pixel color consistency, commonly overlooking the modeling of geometric structures. Consequently, obvious discontinuities such as artifacts and holes appear under novel viewpoints. This discontinuity phenomenon would be more prominent on AD datasets with sparse training views. DRGS \cite{chung2024depth} introduces the dense depth map as geometry guidance to mitigate overfitting.
NeuSG \cite{chen2023neusg} flattens the Gaussian ellipsoid by designing a regularization loss for the scale parameter, and performs consistent joint optimization in combination with the {signed distance function}. 
Scaffold GS \cite{lu2024scaffold} precisely controls the density of the Gaussian ellipsoids by establishing voxel anchors and corresponding offset vectors and learns the various parameters of the Gaussian through a simple multi-layer network. Gaussian Pro \cite{cheng2024gaussianpro} considers the planar prior in the scene, explicitly constraining the growth of Gaussians,
achieving amendatory rendering and more compact representation. 2DGS \cite{huang20242d} presents a highly efficient differentiable 2D Gaussian renderer, enabling perspective-accurate splatting by leveraging 2D surface modeling,
demonstrating superiority in geometric reconstruction.
Driving Gaussians \cite{zhou2024drivinggaussian}, Street Gaussians \cite{yan2024street} and S$^3$ Gaussians \cite{huang2024textit} all initialize the point cloud collected by the LiDAR instead of the point cloud generated by Structure from Motion (SfM). By leveraging dense and accurate data from LiDAR, additional supervision, such as the positional constraint, is introduced to the learning process.

As far as we know, current Gaussian Splatting works for AD in the literature pay little attention to the capability of novel view synthesis, resulting in low-quality images under significant camera transformations. 
We aim to address this issue by decoupling the entire scene into road and non-road regions,
optimizing each with the guidance from hybrid rendering and geometric regularization of Gaussians, achieving more robust and higher-fidelity rendering quality.

\section{Prelimilary}
3DGS \cite{kerbl20233d}  utilizes a set of 3D Gaussians to explicitly represent a scene, achieving high-fidelity rendering quality and fast speed via tile-based rasterization. Each Gaussian $G$ is defined as:
\begin{align}
    G({x})=e^{-\frac12(x-\mu )^\top{\Sigma}^{-1}(x-\mu)},
\end{align}
where $\mu$ is the mean position of Gaussian and $x$ is a discretional coordinate within the Gaussian space. ${\Sigma}=RSS^{T}R^{T}$, where $R$ is the rotation matrix and $S$ is the scaling matrix. Apart from the spherical harmonic coefficients that represent the anisotropic color of each Gaussian, the color of each pixel $\mathbf{x}$ is aggregated by $\alpha$-blending:
\begin{align}
    \mathbf{c}(\mathbf{x})=\sum_{i=1}^{N}{c}_i\alpha_i\prod_{j=1}^{i-1}(1-\alpha_j),
\end{align}
where $\alpha_i$ is the opacity of the 2D Gaussian projected from 3D Gaussian to the 2D image plane and $c_i$ is the color of the Gaussian in the view direction. 

2DGS \cite{huang20242d} ensures multi-view consistency well by adopting the expression of surfels and using the projection method specifically for surfels. It achieves 2D Gaussian rendering through ray-splat intersection. The process of ray-splat intersection can be written as:
\begin{align}
u(\mathbf{x})=\frac{\mathbf{h}_u^2\mathbf{h}_v^4-\mathbf{h}_u^4\mathbf{h}_v^2}{\mathbf{h}_u^1\mathbf{h}_v^2-\mathbf{h}_u^2\mathbf{h}_v^1},\quad v(\mathbf{x})=\frac{\mathbf{h}_u^4\mathbf{h}_v^1-\mathbf{h}_u^1\mathbf{h}_v^4}{\mathbf{h}_u^1\mathbf{h}_v^2-\mathbf{h}_u^2\mathbf{h}_v^1},
\end{align}
where $\mathbf{h}_u$, $\mathbf{h}_v$ are two planes represented in homogeneous coordinates. For the point $\mathbf{u} = (u, v)$,
defining the function $G(\mathbf{u})=\exp\left(-\frac{u^2+v^2}2\right)$, and employing the object-space low-pass filter $\hat{G}(\mathbf{x})=\max\left\{G(\mathbf{u}(\mathbf{x})),G(\frac{\mathbf{x}-\mathbf{c}}{\sigma})\right\}$, the 
rasterization for 2DGS can be shown as:\begin{align}
    \mathbf{c}(\mathbf{x})=\sum_{i=1}{c}_i\alpha_i\hat{G}_i(\mathbf{u}(\mathbf{x}))\prod_{j=1}^{i-1}(1-\alpha_j\hat{G}_j(\mathbf{u}(\mathbf{x}))).
\end{align}

\section{Method}

\subsection{Overview}

We illustrate the overview of the framework in Figure \ref{fig:pipeline}. The proposed method utilizes initial points clouds and semantic masks as paratactic inputs of multi-camera views. The pcd initialization phase first generates road and non-road point clouds, which will be further modeled as decoupled road and environment Gaussian models. Based on the known road point cloud, we propose to design a neural implicit road representation using SDF which serves as a prior for the surface training. The SDF-based surface constraint consists of pre-training and offline supervision phases, by taking advantage of the distance and normal characteristics geometrically. We use distinct Gaussians to model road and non-road elements, enhancing the rendering quality during perspective shifts. To achieve that, the depth-ordered hybrid rendering is delicately designed, by which the road surface and non-road areas can be coupled and superimposed consistently and continuously, yielding superior performance compared with SOTA methods that employ individual Gaussian model. 
The rendered image resulting from the hybrid fusion of two models is supervised by ground truth via Gaussian loss and optimized along with the regularization terms. We will sequentially detail these processes in the following sections.

\subsection{Initialization of the Decoupled PCDs}
For a given clip, we opt to utilize point cloud from the LiDAR that exhibits better multi-view consistency
and prior geometric structure compared to the point cloud from SfM.
The initial point cloud is divided into road and non-road parts for the subsequent training. As shown in Figure \ref{fig:pipeline}, based on the calibrated intrinsic and extrinsic parameters, multi-camera images and their masks generated by Mask2Former \cite{cheng2022masked} are utilized to form colored and semantically labeled single-frame point clouds. 
These individual point clouds are then stitched together to construct road and environment point clouds that cover the entire sequence.
\subsection{Pre-trained Surface Based on SDF}
To optimize the geometric structure of the road and maintain continuity when facing obvious changes in perspective, we design a road constraint based on the surface guidance. Utilizing the segregated road point cloud, we pre-train a signed distance function to represent the road surface implicitly.  This approach differs from the joint ray sampling optimization in NeRF and the guidance of geometric structure and image consistency in GSDF \cite{yu2024gsdf}. As road point clouds collected by LiDAR have been calibrated precisely in advance, the excellent geometric structure is suitable to be the prior for road learning. A straightforward strategy is directly applying distance constraints to Gaussians that belong to the road. Whereas, considering the spatial geometric characteristics of ellipsoids, imposing constraints only on the centers of Gaussian ellipsoids cannot fully restore their correct geometric shapes, potentially restricting the distribution of Gaussian to some extent. We therefore adopt a pre-trained model to optimize the distribution of Gaussian by adjusting Gaussian parameters nearing the road surface. To this end, we design specific regularization terms for road-related Gaussians. These include distance and normal direction constraints that ensure Gaussians stay close to the road surface and align with the normal direction of the road surface concurrently.

We utilize a predefined network $f_{\theta}(x)$ for predicting the corresponding SDF value given the positional coordinate of the point cloud. Data normalization is imposed on all the training and testing coordinates to ensure stability and elevated performance. To equalize the number of points on and off the isosurface, 
we define the road point cloud as $\mathcal{P}_r$, and assign a hypothetical SDF value of zero to each point $x_p \in \mathcal{P}_r$ on the isosurface. For points off the isosurface, we establish a proximity range around each point in $\mathcal{P}_r$, and perform random sampling within this range to obtain sample points  $\mathcal{P}_s\supset \mathcal{P}_r$. 
 For $\forall x_s \in \mathcal{P}_s$, we build a map from point $x_s$ to the signed euclidean distance $d_s$ as:
\begin{align}
    d_s = \text{sgn}(x_s^{(3)}-{x_r^s}^{(3)})\cdot \|x_s - x_r^s\|_2,
\end{align}
where $x_r^s$ is the nearest point of $x_s$ in $\mathcal{P}_r$. Considering that bumps exist in uneven road surfaces, we generate the normal direction for each point in $\mathcal{P}_r$ to assist in the training of the SDF. To be specific,
for $\forall x_r \in \mathcal{P}_r$, set $X_r$ as $k$-nearest neighbors of $x_r$, by performing Singular Value Decomposition (SVD):
\begin{align}
    U\Lambda V^T = \text{svd}(\tilde{X}),
\end{align}
where $\tilde{X} \triangleq X_r - \overline{X}_r$. The right singular vector $v_i$ corresponding to the minimum singular value $\lambda_i$ is finally regarded as the normal of $x_r$, noted as $n_r$.

We supervise the predicted SDF value from the network with the pre-computed ground truth by incorporating a newly proposed normal loss based on the predicted normal direction. The optimization goal for road SDF is written as:
\begin{equation}
\begin{aligned}
\min_{\theta} \big{\{}{L}(\theta)&:= \dfrac{1}{|\mathcal{P}_s|}\sum_{x_s \in \mathcal{P}_s} \|f_{\theta}(\tilde{x}_s) -d_s\|_2^2\\&+ \dfrac{1}{|\mathcal{P}_r|}  \sum_{x_r \in \mathcal{P}_r}\{ \lambda_n\sin^2\langle\nabla f_{\theta}(\tilde{x}_r), n_r\rangle\\
&+\lambda_\text{eik} (\|\nabla f_{\theta}(\tilde{x_r})\|_2-1)^2 \}\big{\}},
\end{aligned}
\end{equation}
in which the first and last terms refer to SDF value loss and eikonal loss respectively, the middle term represents the normal loss, and $|\cdot |$ refers to the number of points in the point cloud. The optimal parameters $\theta^* = \arg\min_{\theta} {L(\theta)}$ will be frozen and used to guide the learning of road Gaussians in the following processes.

\begin{table*}
\centering
\begin{tblr}{
  row{1} = {c},
  row{3} = {c},
  cell{1}{2} = {c=3}{},
  cell{1}{5} = {c=3}{},
  cell{1}{8} = {c=4}{},
  cell{2}{1} = {r=2}{c},
  cell{2}{2} = {r=2}{},
  cell{2}{3} = {r=2}{},
  cell{2}{4} = {r=2}{},
  cell{2}{5} = {r=2}{},
  cell{2}{6} = {r=2}{},
  cell{2}{7} = {r=2}{},
  cell{2}{8} = {c=4}{c},
  cell{4}{1} = {c},
  cell{5}{1} = {c},
  cell{6}{1} = {c},
  cell{7}{1} = {c},
  cell{8}{1} = {c},
  vline{2} = {4-8}{},
  hline{1-2,4,9} = {-}{},
  hline{3} = {8-11}{},
}
\textit{}    & Scene Reconstruction &                &                   & Novel View Synthesis &                &                   & {Free-view Novel View Synthesis}      &        &        &        \\
Metrics      & PSNR$\uparrow$       & SSIM$\uparrow$ & LPIPS$\downarrow$ & PSNR$\uparrow$       & SSIM$\uparrow$ & LPIPS$\downarrow$ & FID$\downarrow$ &        &        &        \\
             &                      &                &                   &                      &                &                   & \textit{Set1}                             & \textit{Set2}   & \textit{Set3}   & \textit{Set4}   \\
3DGS         & 26.91                & 0.8295         & 0.3101            & 25.92                & 0.8089         & 0.3218            & 53.12                            & 61.93  & 82.52  & 61.53  \\
2DGS         & 24.75                & 0.7919         & 0.3746            & 24.26                & 0.7819         & 0.3798            & 101.9                           & 105.5 & 123.4 & 105.6 \\
Gaussian Pro & 25.90                & 0.8019         & 0.3250            & 25.09                & 0.7827         & 0.3358            & 60.52                            & 68.75  & 90.12  & 69.05  \\
Scaffold GS  & \underline{28.04}                & \underline{0.8428}         & \underline{0.3039}            & \textbf{26.87}                & \textbf{0.8224}         & \underline{0.3142}            & \underline{51.93}                    & \underline{58.25}  & \underline{78.64}  & \underline{57.85}  \\
Ours & \textbf{28.09} & \textbf{0.8460} & \textbf{0.2960} &  \underline{26.77} &     \underline{0.8216}  & \textbf{0.3100} & \textbf{48.05} & \textbf{56.93}  & \textbf{75.72}  & \textbf{55.52}     
\end{tblr}
\caption{The quantitative comparison between our method with four SOTA works on PSNR, SSIM, LPIPS, and FID metrics computed over all clips on the Waymo dataset. 
We adopt the same training data for each method and follow the corresponding experimental settings strictly. The details of \textit{Set1} to \textit{Set4} are described in the appendix. The best and second best are marked in \textbf{bold} and \underline{underlined} for each metric respectively.}
\label{tab:main-res}
\end{table*}

\begin{figure}[t]
\centering
\includegraphics[width=0.99\columnwidth]{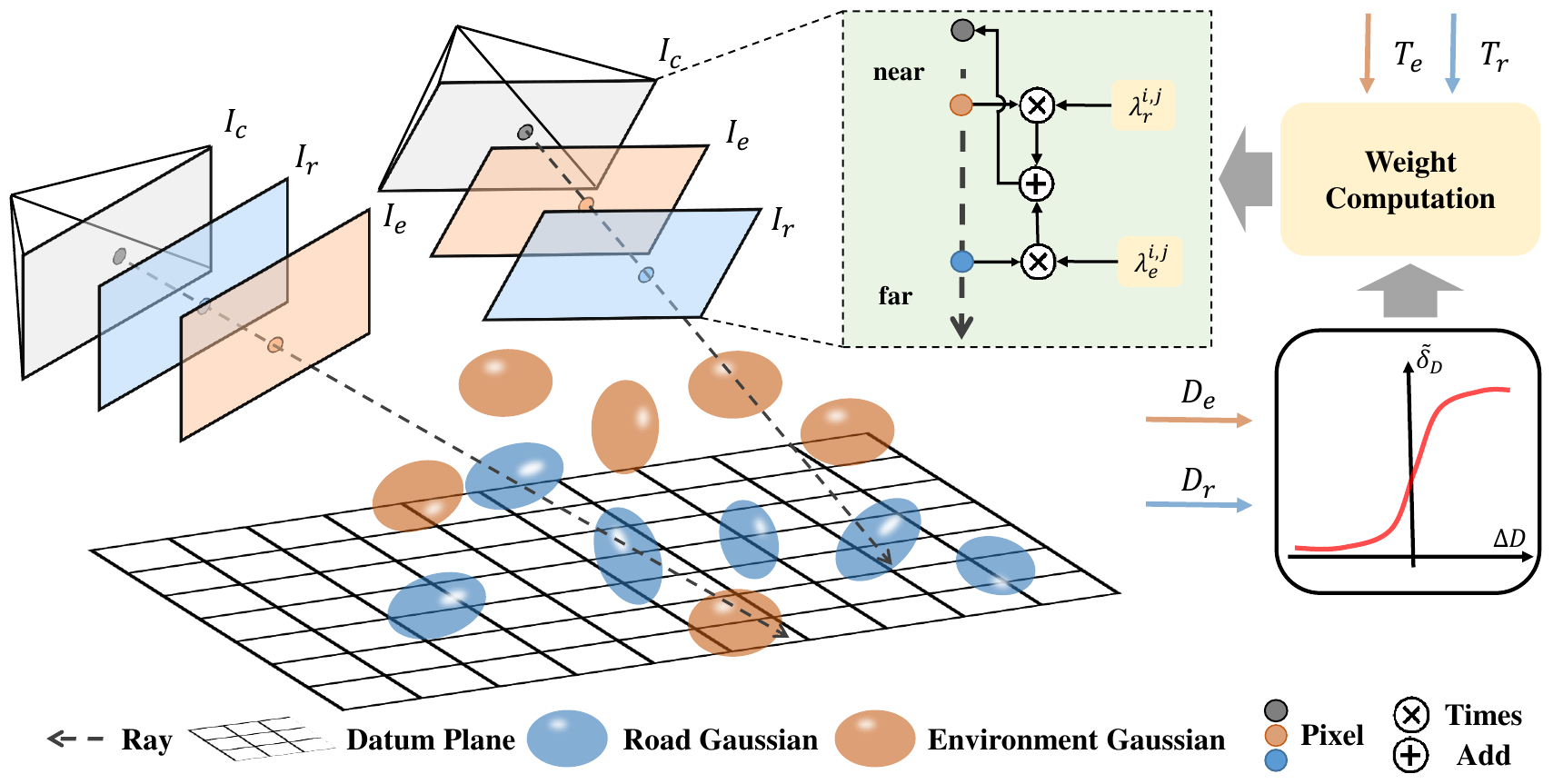} 
\caption{The diagram illustrates the proposed depth-ordered hybrid rendering strategy for the environment and road model. Corresponding primitives of each model generate pixels with colors independently through Gaussian splatting. These colors are then composited based on their rendered depths and transmittances, producing the final rendered image.}
\vspace{-10pt}
\label{fig:depth&order}
\end{figure}
\subsection{Depth-ordered Hybrid Rendering}
Based on the predefined initial semantic non-road point cloud $\mathcal{P}_e$ and road point cloud $\mathcal{P}_r$, we build the environment model and road model simultaneously. In this paper, Scaffold GS and 2DGS are chosen as the fundamental Gaussian splatting substrates. {By executing forward rendering separately, output images $I_{e}$ and $I_{r}$ are generated separately, along with depth maps and accumulated transmittance maps simultaneously.} The depth map $D$ is rendered based on $d_i$ that indicates each splat from the camera, following the rasterization pipeline in \cite{dai2024high, chung2024depth}. We give the definitions of $D$ and $T$ as:
\begin{align}
D = \sum_{i=1}^N d_i \alpha_i \prod_{j=1}^{i-1}(1-\alpha_j), 
T = \prod_{i=1}^{N}(1-\alpha_i).
\end{align}
In our experiments, linearly fusing $I_{e}$
and $I_{r}$ results in inferior image quality. The reason is that the simplistic fusion overlooks the learning of Gaussian properties and the actual depth relationships of Gaussians derived from the environment and road model. Accordingly, we adopt a pixel-level hybrid rendering method based on depth sorting, as depicted in Figure \ref{fig:depth&order}. 
Assuming that the tuple $[I_{e}, D_{e}, T_{e}]$ severally represents the image, depth, and accumulated transmittance rendered by the environment model,  $[I_{r}, D_{r}, T_{r}]$ indicates the image, depth, and accumulated transmittance rendered by the road model in the same manner.
The composite rendering strategy is presented as: 
\begin{align}
    I_c = \lambda_{r} I_{r}+ \lambda_{e}I_{e},
\end{align}
where $\lambda_{r}$ and $\lambda_{e}$ are the weights for fusing $I_{r}$ and $I_{e}$, which are organized as: 

\begin{equation}
\begin{aligned}
\lambda_{r} &= T_{e}\delta_D + (1-\delta_D),\\
\lambda_{e} &= T_{r} (1-\delta_D)+\delta_D,
\end{aligned}
\end{equation}
where $\delta_D$ is the item for composing weights of fusion, which is defined as:
\begin{equation}
\begin{matrix}
\delta_D = \begin{cases}
1 \hspace{2mm}& D_{r}>D_{e},\\
0 & D_{r}\leq D_{e}.
\end{cases}
\end{matrix}  
\label{delta}
\end{equation}
One limitation of blending the rendered images discontinuously is that the weights also vary discretely, which results in the distinct appearance of boundary lines on the edge of road and non-road regions. We thus introduce a consecutive strategy to model the dynamic weights based on the rendered depth. Concretely, the sigmoid function $S(x)$ is utilized to realize the ordered rendering as: 
\begin{equation}
      S(x) = \frac{1}{1+\exp(-s_{\sigma} x)},  
\end{equation}
where $s_\sigma$ is a given hyperparameter.
We set $\Delta D =D_{r}-D_{e}$ indicating depth difference as the input of $S(x)$ to obtain $\tilde{\delta}_D = S(\Delta D)$, the smoother weights can be shown as:

\begin{equation}
\begin{aligned}
\lambda_{r} &= T_{e}\tilde{\delta}_D+ (1-\tilde{\delta}_D),\\
\lambda_{e} &= T_{r} (1-\tilde{\delta}_D)+\tilde{\delta}_D.
\end{aligned}
\end{equation}
With the proposed smooth rendering mechanism, the overall rendering quality can be considerably improved, and the synthesis of new perspectives will appear more natural and fluid. 
\begin{figure*}[ht]
\centering
\includegraphics[width=1.99\columnwidth]{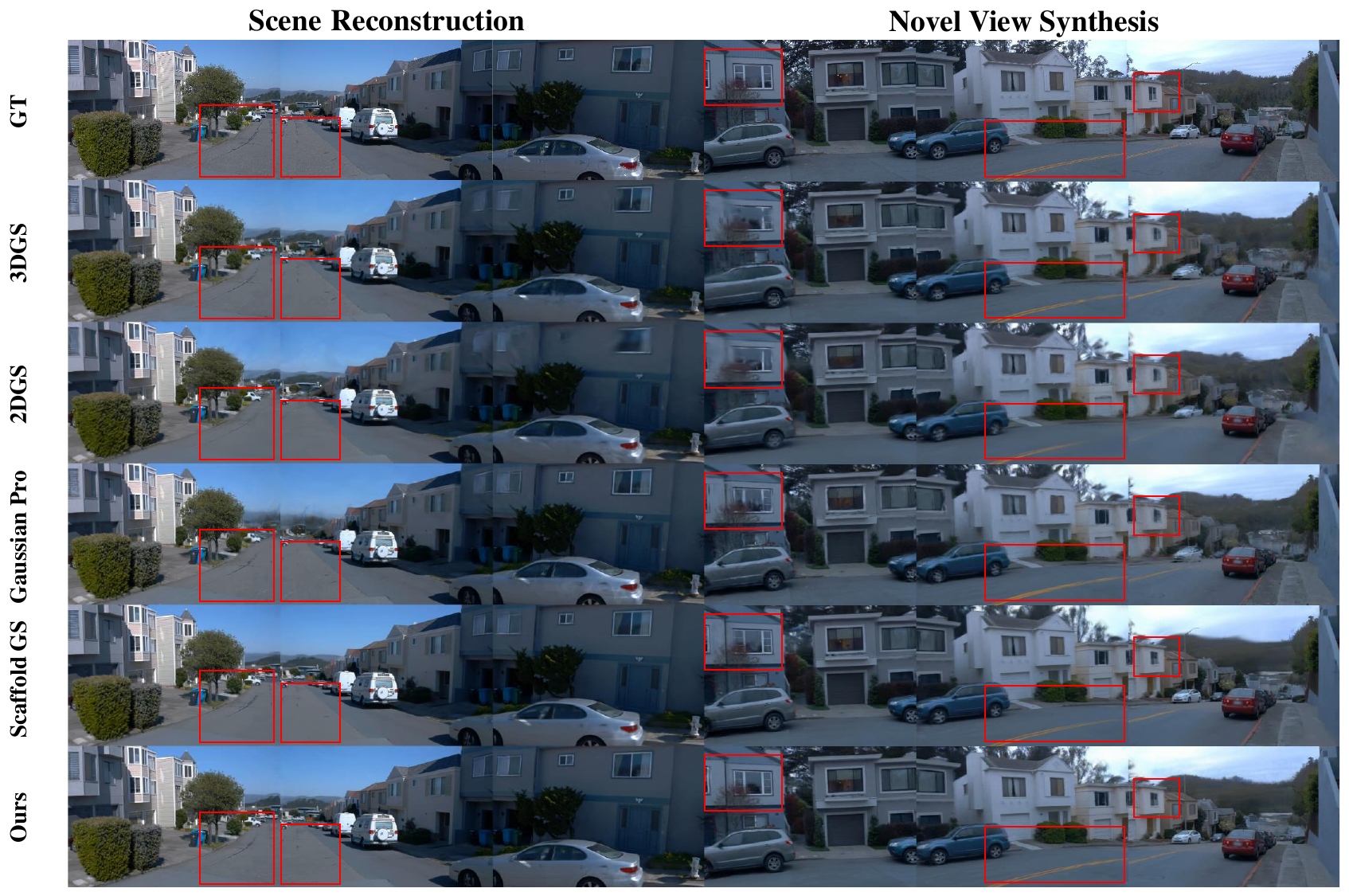} 
\caption{Comparison of different methods on the Waymo dataset, the left column and right column display the quality of scene reconstruction and novel view synthesis respectively. Our method achieves high-quality reconstruction for both the environment and the road areas, excelling over other comparative methods.}
\label{fig:main-rebuild&nvs}
\end{figure*}
\subsection{Loss Functions}
The overall training objective is given as follows:
\begin{align}
\mathcal{L}=\mathcal{L}_\text{gs}\!+\!\lambda_\text{tran}\mathcal{L}_\text{tran}\!+\!\lambda_\text{sdf}\mathcal{L}_\text{sdf} \!+ \!\lambda_\text{cons}\mathcal{L}_\text{cons}\!+\! \lambda_{\text{tv}} \mathcal{L}_{\text{tv}},
\end{align}
among which $\mathcal{L}_\text{gs}$ is responsible for measuring the reconstruction discrepancy used in 3DGS:
\begin{align}
    \mathcal{L}_\text{gs} = \lambda \mathcal{L}_1 (I_c,I_{gt}) +(1-\lambda)\mathcal{L}_\text{D-SSIM}(I_c,I_{gt}),
\end{align}
where $I_{gt}$ is the ground truth.
The $\mathcal{L}_\text{sdf}$ consists of two parts, distance loss and normal loss of the road surface, written as:
\begin{align}
    \mathcal{L}_\text{sdf} \!=\! \dfrac{1}{N}\!\sum_{i=1}^N \! \big{\{}  \lambda_d \|f_{\theta^*}(\tilde{x}_i)\|_1 \!\!+\! \lambda_n\!\sin^2 \langle \nabla f_{\theta^*}\!(\tilde{x}_i), t_n\rangle\! \big{\}},\!
\end{align}
where $t_n = t_u\times t_v$ refers to 2D primitive normal of Gaussian. $\mathcal{L}_\text{tran}$ supervises the accumulated transmittance of road and non-road regions, ensuring their consistency with the corresponding regions in the 2D semantic segmentation, which is represented as:
\begin{align}
    \mathcal{L}_\text{tran} =\dfrac{1}{|M|}\big{(} \|T_{e}- {M}\|_F^2 + \|T_{r}- (1-{M})\|_F^2\big{)},
\end{align}
where ${M}$ refers to the semantic mask of the image, with entries of 1 for road pixels and 0 for non-road pixels. $|\cdot|$ denotes the total number of pixels for a given matrix.
The $\mathcal{L}_\text{cons}$ enforces consistency between the depth rendered by environment Gaussians and road Gaussians:
\begin{align}
    \mathcal{L}_\text{cons} =  \max_{j} \min_{i} |D^{i,j }_{e} - D^{i,j}_{r}|,
\end{align}
where $i$ and $j$ satisfying $\tilde{M}^{i,j}\not=0$, $\tilde{M}$ refers to the banded boundary generated by binary mask $M$. The orange band intuitively indicates the non-zero entries in $\tilde{M}$, as shown in the bottom right of Figure \ref{fig:pipeline}.
We follow Plenoxels \cite{fridovich2022plenoxels} and use a total variation loss
$\mathcal{L}_\text{tv}$.
The cumulative effect of the aforementioned loss functions allows DHGS to produce high-fidelity renderings that maintain a continuous geometric structure.

\section{Experiments and Analysis}

\subsection{Experimental Settings}
\noindent{\textbf{Datasets and metrics.}} We experiment on the publicly available urban scene dataset Waymo \cite{sun2020scalability}, 32 static scenes are adopted as the same as EmerNeRF \cite{yang2023emernerf}. We employ multiple metrics to evaluate the performance of the proposed method and baselines. 
PSNR, SSIM, and LPIPS are utilized to validate the image quality for rendered images with ground truth.
For novel views lacking ground truth, we apply FID score \cite{heusel2017gans} to assess the image quality. The average metrics across all clips are calculated as comparative values.

\noindent{\textbf{Baselines.}} We compare our method with the state-of-the-art and publicly available methods. 3DGS \cite{kerbl20233d}, 2DGS \cite{huang20242d}, Gaussian Pro \cite{cheng2024gaussianpro}, and Scaffold GS \cite{lu2024scaffold} are chosen for comparison, recognized for their exceptional performances.

\begin{figure*}[t]
\centering
\includegraphics[width=1.99\columnwidth]{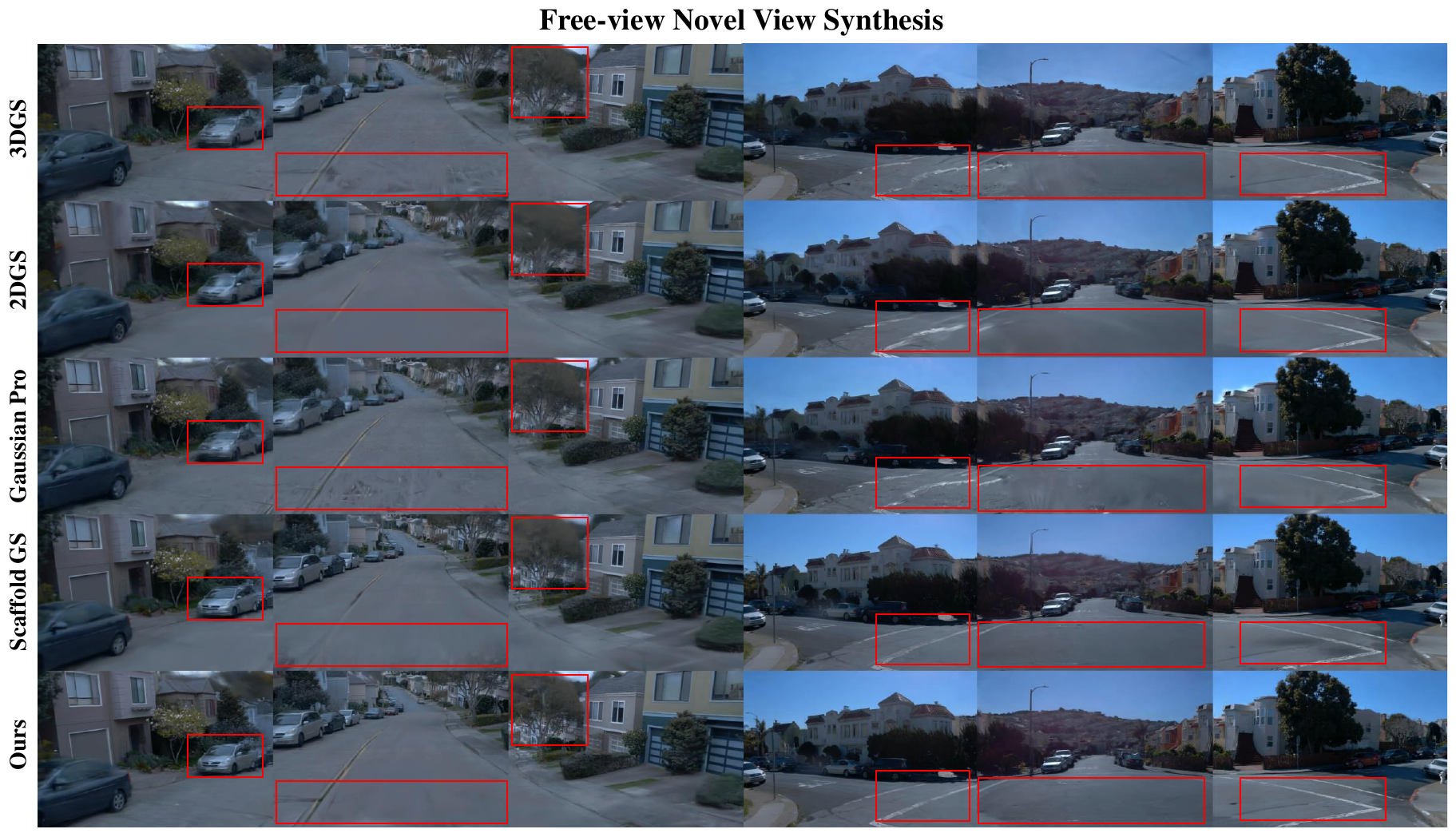} 
\caption{Visual comparisons on the free-view novel view synthesis. The left and right columns respectively exhibit the results of \textit{Set3} and \textit{Set4} under the free viewpoint setting, where our method significantly outperforms other comparative methods in capturing both road and environment details.}
\label{fig:main-free-nvs}
\end{figure*}

\noindent{\textbf{Implementation details.}} 
The initial point cloud is randomly downsampled to 600,000 points. Our method and the other comparison methods are trained for 60,000 iterations. For the pre-trained SDF network, we use an 8-layer MLP, training it for 20,000 iterations. The normals of surface points are derived using SVD with 50 neighborhoods. The weights for the normal and eikonal losses in the SDF network are carefully calibrated, set to 0.1 and 0.01, respectively.
The weights $\lambda_\text{tran}$, $\lambda_\text{sdf}$, $\lambda_\text{cons}$, $\lambda_\text{tv}$, $\lambda_d$ and $\lambda_n$ of the overall training loss are set to $0.1$, $1.0$, $0.04$, $0.1$, $0.1$ and $0.1$. We set the densification intervals for the Gaussian primitives in the environment model and road model to $200$ and $300$ respectively. The learning rate for the anchor position in the environment model is set as $0.00016$, the remaining learning rates are kept consistent with the original settings. For each clip, we utilize all the front, left-front, and right-front views of the entire sequence for training, with a 4:1 split ratio between the training and testing sets. For simplicity, the images in the training set are referred to as the reconstruction views, while those in the testing set are called the novel views.
We further apply unrestricted transformations to camera poses to test the robustness of the model in data-driven simulation, termed as free-view novel view synthesis.

\subsection{Comparisons}
We benchmark our method against established baselines in scene reconstruction, novel view synthesis, and free-view novel view synthesis which signifies notable viewpoint transfer. All methods employ a consistent set of reconstruction views for training, while the remainder are employed as novel views for testing in each clip. 
Qualitative and quantitative results are shown in the following sections, 
including an ablation study for detailed analysis.

\noindent{\textbf{Scene Reconstruction on the Waymo dataset.}} For reconstruction views, we calculate the mean PSNR, SSIM, and LPIPS of the rendered images across 32 clips, as shown in Table \ref{tab:main-res}. Our method outperforms all the comparative methods: including 3DGS, 2DGS, Gaussian Pro, and Scaffold GS. The left part of Figure \ref{fig:main-rebuild&nvs} displays the rendered images of the scene reconstruction from our method and its competitors. It can be noticed that 
our method delivers superior reconstruction quality with finer details, particularly in the road area, where our rendering exhibits richer texture information.

\noindent{\textbf{Novel View Synthesis on the Waymo dataset.}} We assess the aforementioned methods on the novel view synthesis. Table \ref{tab:main-res} illustrates the validation results, demonstrating that our method surpasses most rival methods, such as 3DGS, 2DGS, and Gaussian Pro, and achieves comparable results to Scaffold GS. Notably, 
Scaffold GS leads in PSNR and SSIM metrics. 
As depicted in the right part of Figure \ref{fig:main-rebuild&nvs}, blurriness that is highlighted by red box appears in some non-road areas for 3DGS, 2DGS, and Gaussian Pro. In contrast, our method restores more road details which can be seen in lane lines within the red boxes, owing to the imposition of geometric constraints on the road. Whereas, other methods such as 2DGS, Gaussian Pro, and Scaffold GS fail to preserve these subtleties. Despite novel views not being part of the training set, their camera pose variations are minimal compared to trained views, 
suggesting they can be considered as interpolations of the trained views.
Such minor variations in novel views do not sufficiently reflect the capability of geometric representation of the model.

\noindent{\textbf{Free-view Novel View Synthesis on the Waymo dataset.}} To further underscore the geometric consistency of our approach, we evaluate the trained model on free-view novel views with more pronounced camera pose variations. We adjust the poses of the cameras relative to the vehicle, mimicking various observation perspectives for different vehicle types with diverse camera positions. This paper details four types of camera pose transformations, with specific configurations available in the supplementary material.
Given the absence of ground truth for these novel views, we adopt the FID score, which excels at measuring the feature discrepancy between the original and the free-view novel views. The comparisons of FID for free-view novel views are also presented in Table \ref{tab:main-res}, where our method achieves the highest performance compared to other methods in all four configurations with a significant margin of superiority. The qualitative rendering results, depicted in Figure \ref{fig:main-free-nvs}, reveal that free-view rendered images produced by other methods suffer from extensive artifacts and lane line breakage. In contrast, our method consistently captures environment and road details, including lane lines, even with substantial variations in camera poses. The results further confirm that the regularization constraints and the depth-ordered hybrid rendering strategy we proposed facilitate the acquisition of an optimal geometric structure, thereby showcasing the robustness of our approach in free-view novel view synthesis. 
More visual comparisons can be found in supplemental materials.


\begin{figure}[t]
\centering
\includegraphics[width=0.9\columnwidth]{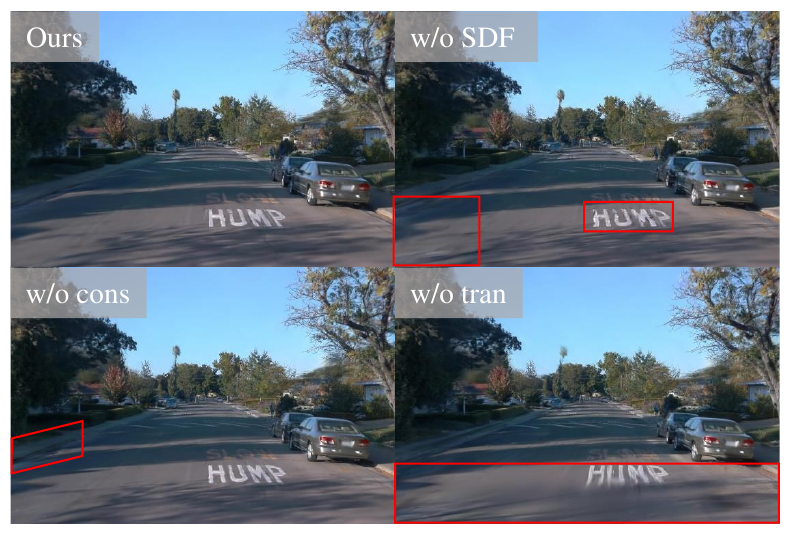} 
\caption{{Qualitative comparison for free-view novel views of ablation studies. 
The emphasized patches reveal that these regularization terms contribute to the accurate reconstruction of road geometry.}}
\label{fig:ablation}
\end{figure}

\subsection{Ablation Study} 
To assess the  efficacy of our proposed regularization components, we perform an ablation study,
systematically omitting the SDF, transmittance and consistency losses, denoted as "w/o SDF", "w/o tran", and "w/o cons".
The quantitative results for scene reconstruction, novel view synthesis, and free-view novel view synthesis are consolidated in Table \ref{tab:ablation}. For scene reconstruction and novel view synthesis, the absence of the SDF loss leads to a decline in performance metrics, whereas the exclusion of transmittance loss or consistency loss has a minimal impact. Conversely, for free-view novel view synthesis, the removal of any loss escalates the FID score, signifying a greater divergence between the novel view and the originals. To more intuitively illustrate the specific contributions of these components, we present the visualization results of free-view novel view synthesis in Figure \ref{fig:ablation}. Specifically, the SDF loss ensures the fidelity of road geometry, while the consistency loss guarantees a smooth integration between the environment and road models. The transmittance loss is crucial for the precise alignment of the rendering regions of both models with the 2D segmentation, thereby averting rendering artifacts. Therefore, each regularization term is integral to our DHGS framework.


\renewcommand{\arraystretch}{1.5}
\begin{table}[t]
\centering
\normalsize
\begin{tabular}{ccccc}
\hline
\multicolumn{1}{c}{\textit{Metrics}} & w/o SDF      &    w/o tran  & w/o cons     & Ours     \\ \hline
                                     & \multicolumn{4}{c}{\textit{Scene Reconstruction}} \\ \hline
\multicolumn{1}{c|}{PSNR$\uparrow$}            &    27.76      &  \textbf{28.18}           &       28.01     &      28.09    \\
\multicolumn{1}{c|}{SSIM$\uparrow$}            &    0.8381          &       \textbf{0.8473}     &      0.8453      &    0.8460      \\
\multicolumn{1}{c|}{LPIPS$\downarrow$}           &        0.3147      &      \textbf{0.2934}      &        0.3007    &      0.2961    \\ \hline
                                     & \multicolumn{4}{c}{\textit{Novel View Synthesis}} \\ \hline
\multicolumn{1}{c|}{PSNR$\uparrow$}            &        26.29      &     26.63       &     26.68       &     \textbf{ 26.77 }   \\
\multicolumn{1}{c|}{SSIM$\uparrow$}            &        0.8151     &    0.8213        &    0.8205        &     \textbf{0.8217}    \\
\multicolumn{1}{c|}{LPIPS$\downarrow$}           &   0.3195           &    0.3113        &         0.3154   &   \textbf{0.3100}      \\ \hline
                                     & \multicolumn{4}{c}{\textit{Free-view Novel View Synthesis}} \\ \hline
\multicolumn{1}{c|}{FID$\downarrow$}             &     61.16         &  61.02          &        62.49    &    \textbf{59.06}      \\ \hline
\end{tabular}
\caption{Quantitative results of ablation experiments on the Waymo dataset, with each metric averaged across all clips.
}
\label{tab:ablation}
\end{table}
\section{Conclusions}
In this paper, we propose a novel reconstruction method called DHGS, which aims to elevate the rendering quality of novel view synthesis for static driving scenes. DHGS leverages two decoupled models to represent the road and environment separately, 
which are then merged via a pixel-level hybrid renderer, facilitated by our innovative depth-ordered hybrid rendering strategy.
Additionally, consistency loss and transmittance loss are proposed to constrain the basal models, enabling the generation of continuously consistent rendered images. Moreover, we develop a road surface regularization term grounded in SDF to ensure better geometric consistency of the road surface, guaranteeing stable image quality as perspective changes drastically. Through extensive testing on the Waymo dataset, 
our method demonstrates the state-of-the-art performance in reconstruction and free-view novel view synthesis.


\bibliography{aaai}

\begin{thebibliography}{}

\bibitem[\protect\citeauthoryear{Barron \bgroup et al\mbox.\egroup }{2021}]{barron2021mip}
Barron, J.~T.; Mildenhall, B.; Tancik, M.; Hedman, P.; Martin-Brualla, R.; and Srinivasan, P.~P.
\newblock 2021.
\newblock Mip-nerf: A multiscale representation for anti-aliasing neural radiance fields.
\newblock In {\em Proceedings of the IEEE/CVF International Conference on Computer Vision (ICCV)},  5855--5864.

\bibitem[\protect\citeauthoryear{Barron \bgroup et al\mbox.\egroup }{2022}]{barron2022mip}
Barron, J.~T.; Mildenhall, B.; Verbin, D.; Srinivasan, P.~P.; and Hedman, P.
\newblock 2022.
\newblock Mip-nerf 360: Unbounded anti-aliased neural radiance fields.
\newblock In {\em Proceedings of the IEEE/CVF Conference on Computer Vision and Pattern Recognition (CVPR)},  5470--5479.

\bibitem[\protect\citeauthoryear{Chen, Li, and Lee}{2023}]{chen2023neusg}
Chen, H.; Li, C.; and Lee, G.~H.
\newblock 2023.
\newblock Neusg: Neural implicit surface reconstruction with 3d gaussian splatting guidance.
\newblock {\em arXiv preprint arXiv:2312.00846}.

\bibitem[\protect\citeauthoryear{Cheng \bgroup et al\mbox.\egroup }{2022}]{cheng2022masked}
Cheng, B.; Misra, I.; Schwing, A.~G.; Kirillov, A.; and Girdhar, R.
\newblock 2022.
\newblock Masked-attention mask transformer for universal image segmentation.
\newblock In {\em Proceedings of the IEEE/CVF Conference on Computer Vision and Pattern recognition (CVPR)},  1290--1299.

\bibitem[\protect\citeauthoryear{Cheng \bgroup et al\mbox.\egroup }{2024}]{cheng2024gaussianpro}
Cheng, K.; Long, X.; Yang, K.; Yao, Y.; Yin, W.; Ma, Y.; Wang, W.; and Chen, X.
\newblock 2024.
\newblock Gaussianpro: 3d gaussian splatting with progressive propagation.
\newblock {\em arXiv preprint arXiv:2402.14650}.

\bibitem[\protect\citeauthoryear{Chung, Oh, and Lee}{2024}]{chung2024depth}
Chung, J.; Oh, J.; and Lee, K.~M.
\newblock 2024.
\newblock Depth-regularized optimization for 3d gaussian splatting in few-shot images.
\newblock In {\em Proceedings of the IEEE/CVF Conference on Computer Vision and Pattern Recognition (CVPR)},  811--820.

\bibitem[\protect\citeauthoryear{Dai \bgroup et al\mbox.\egroup }{2024}]{dai2024high}
Dai, P.; Xu, J.; Xie, W.; Liu, X.; Wang, H.; and Xu, W.
\newblock 2024.
\newblock High-quality surface reconstruction using gaussian surfels.
\newblock {\em arXiv preprint arXiv:2404.17774}.

\bibitem[\protect\citeauthoryear{Fridovich-Keil \bgroup et al\mbox.\egroup }{2022}]{fridovich2022plenoxels}
Fridovich-Keil, S.; Yu, A.; Tancik, M.; Chen, Q.; Recht, B.; and Kanazawa, A.
\newblock 2022.
\newblock Plenoxels: Radiance fields without neural networks.
\newblock In {\em Proceedings of the IEEE/CVF Conference on Computer Vision and Pattern Recognition (CVPR)},  5501--5510.

\bibitem[\protect\citeauthoryear{Guo \bgroup et al\mbox.\egroup }{2023}]{guo2023streetsurf}
Guo, J.; Deng, N.; Li, X.; Bai, Y.; Shi, B.; Wang, C.; Ding, C.; Wang, D.; and Li, Y.
\newblock 2023.
\newblock Streetsurf: Extending multi-view implicit surface reconstruction to street views.
\newblock {\em arXiv preprint arXiv:2306.04988}.

\bibitem[\protect\citeauthoryear{Heusel \bgroup et al\mbox.\egroup }{2017}]{heusel2017gans}
Heusel, M.; Ramsauer, H.; Unterthiner, T.; Nessler, B.; and Hochreiter, S.
\newblock 2017.
\newblock Gans trained by a two time-scale update rule converge to a local nash equilibrium.
\newblock {\em Advances in Neural Information Processing Systems (NIPS)} 30.

\bibitem[\protect\citeauthoryear{Huang \bgroup et al\mbox.\egroup }{2024a}]{huang20242d}
Huang, B.; Yu, Z.; Chen, A.; Geiger, A.; and Gao, S.
\newblock 2024a.
\newblock 2d gaussian splatting for geometrically accurate radiance fields.
\newblock {\em arXiv preprint arXiv:2403.17888}.

\bibitem[\protect\citeauthoryear{Huang \bgroup et al\mbox.\egroup }{2024b}]{huang2024textit}
Huang, N.; Wei, X.; Zheng, W.; An, P.; Lu, M.; Zhan, W.; Tomizuka, M.; Keutzer, K.; and Zhang, S.
\newblock 2024b.
\newblock S$^3$ gaussian: Self-supervised street gaussians for autonomous driving.
\newblock {\em arXiv preprint arXiv:2405.20323}.

\bibitem[\protect\citeauthoryear{Kerbl \bgroup et al\mbox.\egroup }{2023}]{kerbl20233d}
Kerbl, B.; Kopanas, G.; Leimk{\"u}hler, T.; and Drettakis, G.
\newblock 2023.
\newblock 3d gaussian splatting for real-time radiance field rendering.
\newblock {\em ACM Transactions on Graphics (TOG)} 42(4):1--14.

\bibitem[\protect\citeauthoryear{Li \bgroup et al\mbox.\egroup }{2024}]{li2023lanesegnet}
Li, T.; Jia, P.; Wang, B.; Chen, L.; Jiang, K.; Yan, J.; and Li, H.
\newblock 2024.
\newblock Lanesegnet: Map learning with lane segment perception for autonomous driving.
\newblock In {\em Proceedings of the International Conference on Learning Representations (ICLR)}.

\bibitem[\protect\citeauthoryear{Lindstr{\"o}m \bgroup et al\mbox.\egroup }{2024}]{lindstrom2024nerfs}
Lindstr{\"o}m, C.; Hess, G.; Lilja, A.; Fatemi, M.; Hammarstrand, L.; Petersson, C.; and Svensson, L.
\newblock 2024.
\newblock Are nerfs ready for autonomous driving? towards closing the real-to-simulation gap.
\newblock In {\em Proceedings of the IEEE/CVF Conference on Computer Vision and Pattern Recognition (CVPR)},  4461--4471.

\bibitem[\protect\citeauthoryear{Lu \bgroup et al\mbox.\egroup }{2024}]{lu2024scaffold}
Lu, T.; Yu, M.; Xu, L.; Xiangli, Y.; Wang, L.; Lin, D.; and Dai, B.
\newblock 2024.
\newblock Scaffold-gs: Structured 3d gaussians for view-adaptive rendering.
\newblock In {\em Proceedings of the IEEE/CVF Conference on Computer Vision and Pattern Recognition (CVPR)},  20654--20664.

\bibitem[\protect\citeauthoryear{Miao \bgroup et al\mbox.\egroup }{2024}]{EDUS}
Miao, S.; Huang, J.; Bai, D.; Qiu, W.; Liu, B.; Geiger, A.; and Liao, Y.
\newblock 2024.
\newblock Edus: Efficient depth-guided urban view synthesis.
\newblock In {\em European Conference on Computer Vision (ECCV)}.

\bibitem[\protect\citeauthoryear{Mildenhall \bgroup et al\mbox.\egroup }{2020}]{mildenhall2020nerf}
Mildenhall, B.; Srinivasan, P.; Tancik, M.; Barron, J.; Ramamoorthi, R.; and Ng, R.
\newblock 2020.
\newblock Nerf: Representing scenes as neural radiance fields for view synthesis.
\newblock In {\em European Conference on Computer Vision (ECCV)}.

\bibitem[\protect\citeauthoryear{M{\"u}ller \bgroup et al\mbox.\egroup }{2022}]{muller2022instant}
M{\"u}ller, T.; Evans, A.; Schied, C.; and Keller, A.
\newblock 2022.
\newblock Instant neural graphics primitives with a multiresolution hash encoding.
\newblock {\em ACM Transactions on Graphics (TOG)} 41(4):1--15.

\bibitem[\protect\citeauthoryear{Rudin and Osher}{1994}]{rudin1994total}
Rudin, L.~I., and Osher, S.
\newblock 1994.
\newblock Total variation based image restoration with free local constraints.
\newblock In {\em Proceedings of 1st International Conference on Image Processing (ICIP)}, volume~1,  31--35.
\newblock IEEE.

\bibitem[\protect\citeauthoryear{Sun \bgroup et al\mbox.\egroup }{2020}]{sun2020scalability}
Sun, P.; Kretzschmar, H.; Dotiwalla, X.; Chouard, A.; Patnaik, V.; Tsui, P.; Guo, J.; Zhou, Y.; Chai, Y.; Caine, B.; et~al.
\newblock 2020.
\newblock Scalability in perception for autonomous driving: Waymo open dataset.
\newblock In {\em Proceedings of the IEEE/CVF Conference on Computer Vision and Pattern Recognition (CVPR)},  2446--2454.

\bibitem[\protect\citeauthoryear{Tonderski \bgroup et al\mbox.\egroup }{2024}]{tonderski2024neurad}
Tonderski, A.; Lindstr{\"o}m, C.; Hess, G.; Ljungbergh, W.; Svensson, L.; and Petersson, C.
\newblock 2024.
\newblock Neurad: Neural rendering for autonomous driving.
\newblock In {\em Proceedings of the IEEE/CVF Conference on Computer Vision and Pattern Recognition (CVPR)},  14895--14904.

\bibitem[\protect\citeauthoryear{Wang \bgroup et al\mbox.\egroup }{2023}]{wang2023f2}
Wang, P.; Liu, Y.; Chen, Z.; Liu, L.; Liu, Z.; Komura, T.; Theobalt, C.; and Wang, W.
\newblock 2023.
\newblock F2-nerf: Fast neural radiance field training with free camera trajectories.
\newblock In {\em Proceedings of the IEEE/CVF Conference on Computer Vision and Pattern Recognition (CVPR)},  4150--4159.

\bibitem[\protect\citeauthoryear{Wu \bgroup et al\mbox.\egroup }{2023}]{wu2023mars}
Wu, Z.; Liu, T.; Luo, L.; Zhong, Z.; Chen, J.; Xiao, H.; Hou, C.; Lou, H.; Chen, Y.; Yang, R.; et~al.
\newblock 2023.
\newblock Mars: An instance-aware, modular and realistic simulator for autonomous driving.
\newblock In {\em CAAI International Conference on Artificial Intelligence},  3--15.
\newblock Springer.

\bibitem[\protect\citeauthoryear{Yan \bgroup et al\mbox.\egroup }{2024}]{yan2024street}
Yan, Y.; Lin, H.; Zhou, C.; Wang, W.; Sun, H.; Zhan, K.; Lang, X.; Zhou, X.; and Peng, S.
\newblock 2024.
\newblock Street gaussians for modeling dynamic urban scenes.
\newblock {\em arXiv preprint arXiv:2401.01339}.

\bibitem[\protect\citeauthoryear{Yang \bgroup et al\mbox.\egroup }{2023a}]{yang2023emernerf}
Yang, J.; Ivanovic, B.; Litany, O.; Weng, X.; Kim, S.~W.; Li, B.; Che, T.; Xu, D.; Fidler, S.; Pavone, M.; et~al.
\newblock 2023a.
\newblock Emernerf: Emergent spatial-temporal scene decomposition via self-supervision.
\newblock {\em arXiv preprint arXiv:2311.02077}.

\bibitem[\protect\citeauthoryear{Yang \bgroup et al\mbox.\egroup }{2023b}]{yang2023unisim}
Yang, Z.; Chen, Y.; Wang, J.; Manivasagam, S.; Ma, W.-C.; Yang, A.~J.; and Urtasun, R.
\newblock 2023b.
\newblock Unisim: A neural closed-loop sensor simulator.
\newblock In {\em Proceedings of the IEEE/CVF Conference on Computer Vision and Pattern Recognition (CVPR)},  1389--1399.

\bibitem[\protect\citeauthoryear{Yu \bgroup et al\mbox.\egroup }{2024}]{yu2024gsdf}
Yu, M.; Lu, T.; Xu, L.; Jiang, L.; Xiangli, Y.; and Dai, B.
\newblock 2024.
\newblock Gsdf: 3dgs meets sdf for improved rendering and reconstruction.
\newblock {\em arXiv preprint arXiv:2403.16964}.

\bibitem[\protect\citeauthoryear{Zhang \bgroup et al\mbox.\egroup }{2020}]{zhang2020nerf++}
Zhang, K.; Riegler, G.; Snavely, N.; and Koltun, V.
\newblock 2020.
\newblock Nerf++: Analyzing and improving neural radiance fields.
\newblock {\em arXiv preprint arXiv:2010.07492}.

\bibitem[\protect\citeauthoryear{Zhou \bgroup et al\mbox.\egroup }{2024a}]{zhou2024hugs}
Zhou, H.; Shao, J.; Xu, L.; Bai, D.; Qiu, W.; Liu, B.; Wang, Y.; Geiger, A.; and Liao, Y.
\newblock 2024a.
\newblock Hugs: Holistic urban 3d scene understanding via gaussian splatting.
\newblock In {\em Proceedings of the IEEE/CVF Conference on Computer Vision and Pattern Recognition (CVPR)},  21336--21345.

\bibitem[\protect\citeauthoryear{Zhou \bgroup et al\mbox.\egroup }{2024b}]{zhou2024drivinggaussian}
Zhou, X.; Lin, Z.; Shan, X.; Wang, Y.; Sun, D.; and Yang, M.-H.
\newblock 2024b.
\newblock Drivinggaussian: Composite gaussian splatting for surrounding dynamic autonomous driving scenes.
\newblock In {\em Proceedings of the IEEE/CVF Conference on Computer Vision and Pattern Recognition (CVPR)},  21634--21643.

\end{thebibliography}

\clearpage
\appendix
\setcounter{figure}{0}
\setcounter{table}{0}
\renewcommand{\thefigure}{A\arabic{figure}}
\renewcommand{\thetable}{A\arabic{table}}

\section{Supplemental Material for DHGS: Decoupled Hybrid Gaussian Splatting for Driving Scene}

In the supplementary material, we introduce more details on the method and implementation in Sec. A. Extra experimental results are illustrated in Sec. B.   

\subsection{A. Implementation Details}
\begin{figure}[thbp]
\centering
\includegraphics[width=0.85\columnwidth]{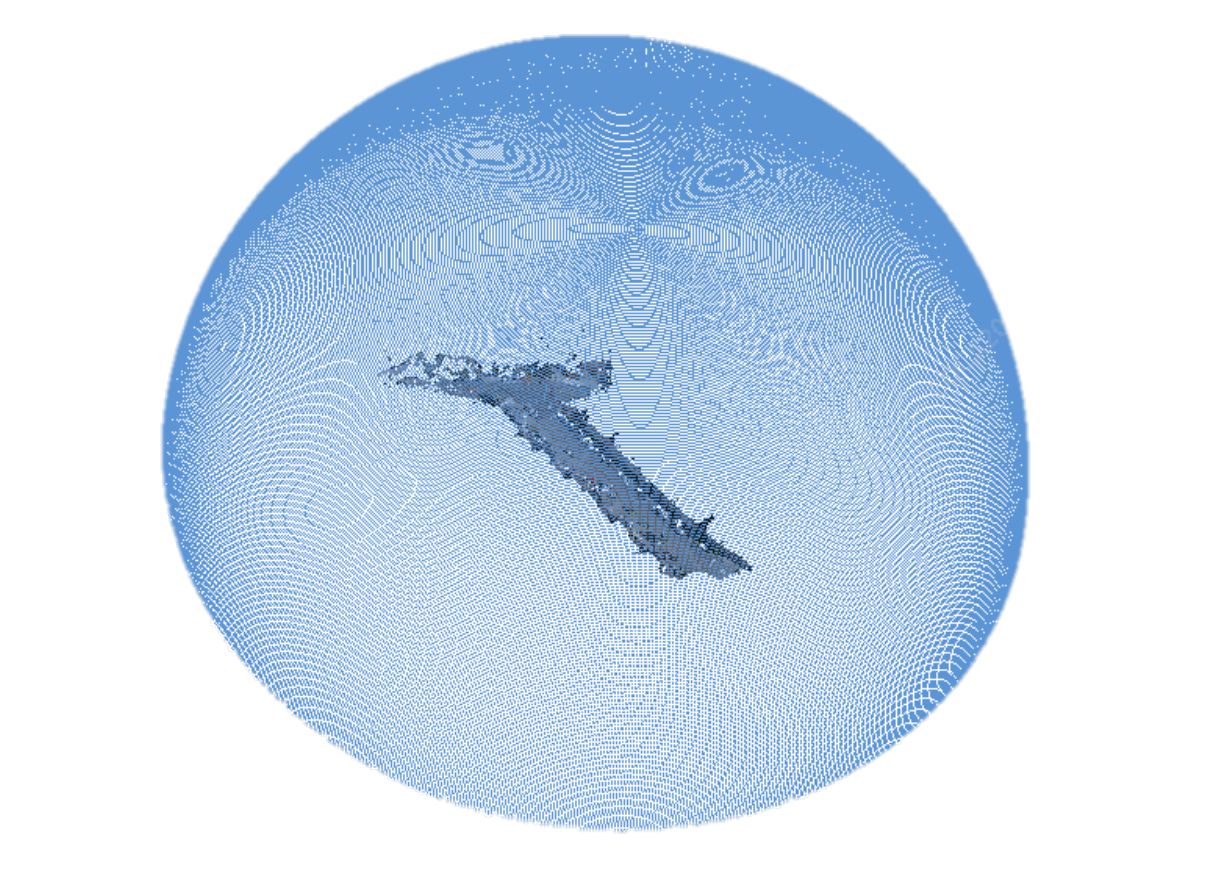} 
\caption{ Visualization of incorporating a sky sphere point cloud into the initial point cloud.}
\vspace{-10pt}
\label{fig:add-sky}
\end{figure}

\paragraph{Initial PCD with sky sphere.}
To address the floaters phenomenon in new perspectives, we devise a straightforward yet effective strategy. We construct a hemispherical virtual sky point cloud layer that envelops the LiDAR point cloud. The radius and center of this sky point cloud are determined by the scale and the average coordinates of the LiDAR points, ensuring an accurate representation of the spatial properties within the sky. Additionally, we apply appropriate coloring to the sky point cloud to maintain its visual integrity, as depicted in Figure \ref{fig:add-sky}. Table \ref{tab:add-sky} presents a comparative analysis between incorporating the sky sphere point cloud into 3DGS and not incorporating it. It is evident that the introduction of sky sphere pcd is able to improve the rendering quality, no matter in reconstruction or novel view synthesis. For fairness, we integrate this sky sphere point cloud mechanism into all comparative methods, such that the potential biases stemming from differences in sky handling can be eliminated.


\renewcommand{\arraystretch}{1.5}
\begin{table}[ht]
\small
\centering
\begin{tblr}{
  cells = {c},
  vline{2} = {2-3}{},
  hline{1-2,4} = {-}{},
}
{Initial PCD}         & PSNR$\uparrow$  & SSIM$\uparrow$  & LPIPS$\downarrow$  & FID$\downarrow$ \\
PCD w/o sky sphere &    25.63  &   0.8062   &  0.3265  &  67.52 \\
PCD w/ sky sphere &    \textbf{25.92}  &   \textbf{0.8089}   &  \textbf{0.3218}  &  \textbf{64.77}
\end{tblr}
\caption{The quantitative results of 3DGS with and without incorporating the sky sphere point cloud into the initial point cloud. The FID here is the average based on the FIDs of all four sets.}
\label{tab:add-sky}
\vspace{-15pt}
\end{table}
\paragraph{Sigmoid distance constraint.} 
In addition to optimizing the road model, we also utilize a distance constraint mechanism for the environment model. Rather than directly applying distance regularization to the offsets, we adopt a more restrictive approach, overlaying a sigmoid function on the offsets. 
This approach ensures more stringent control over the boundaries.
Specifically, we implement the transformation using the following formula: 
\[
\mu_k = x_v + \lambda_{\sigma}\cdot\big{(}2\cdot \text{sigmoid}(\mathcal{O}_v\cdot l_v)-1\big{)},
\] 
where $x_v$ refers to the coordinate of the anchor point, $\mu_k$ refers to the center coordinates of the $k$-th neural Gaussian corresponding to the anchor. $l_v$ and $\mathcal{O}_v$ represent the scaling factor and offset vector respectively.
This design ensures that the offsets are strictly confined within a reasonable and finite range, thereby preventing the generation of neural Gaussian ellipsoids that are excessively distant from their anchor points. Consequently, the power of the environment model to represent fine details is significantly enhanced.
\begin{figure}[htbp]
\centering
\includegraphics[width=0.99\columnwidth]{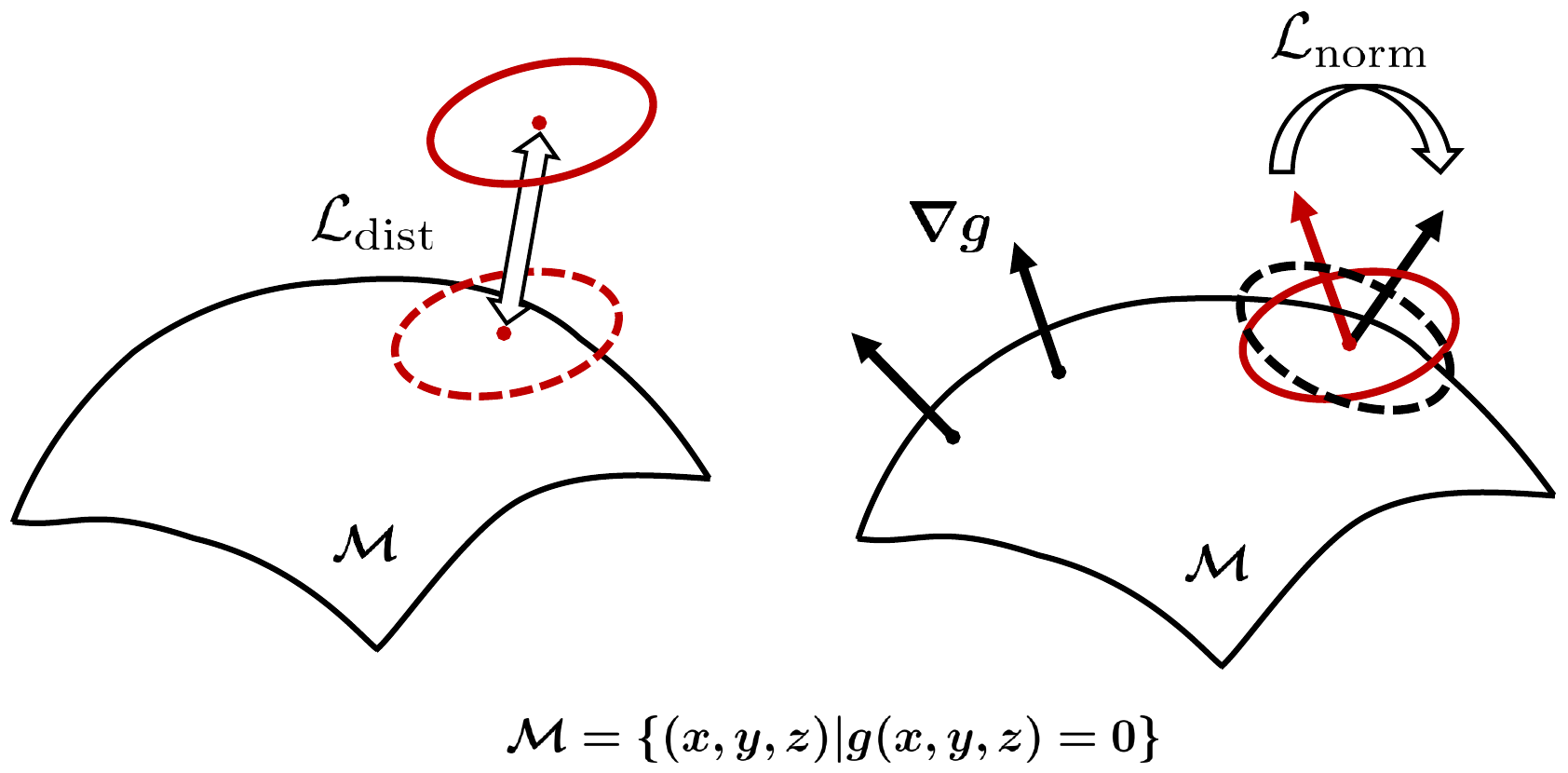} 
\caption{The simplified sketch of the SDF regularizer, where  $\mathcal{L}_\text{dist}$ and $\mathcal{L}_\text{norm}$  refer to the distance constraint and normal constraint  guided by the zero-level set of SDF, the function $g$ refers to the ideal pretrained SDF.}
\label{fig:dist&norm-cons}
\vspace{-15pt}
\end{figure}

\paragraph{Regularization loss.} The total variation regularization $\mathcal{L}_\text{tv}$ \cite{rudin1994total} is applied in this paper to refine the rendered pixel depths. The depth map can be generated through a rendering process similar to the one we calculate the depth map: $ D = \lambda_{r} D_{r}+\lambda_{e} D_{e}$, to maintain the continuity of $D$ near the boundaries of two depth maps, the TV regularizer can be represented as: 
\[
\mathcal{L}_\text{tv} = \sum_{\tilde{M}^{i,j} \not=0}\big{(}(D^{i,j-1}-D^{i,j})^2 +(D^{i+1,j}-D^{i,j})^2 \big{)}^{\frac{1}{2}}.
\]
The SDF regularizer $\mathcal{L}_\text{sdf}$, inspired by NeuSG \cite{chen2023neusg} originally optimizing the positions and orientations of 3D Gaussians, is applied in this paper similarly to 2D Gaussians. Specifically, we leverage the gradient direction of the pre-trained SDF to guide the orientation updates of the 2D Gaussians, while simultaneously constraining their disk centers to align with the SDF. This approach guarantees a seamless integration, which is intuitively illustrated in Figure \ref{fig:dist&norm-cons}.

\begin{table*}[htbp]
\centering
\begin{tblr}{
  cells = {c},
  cell{1}{1} = {r=2}{},
  cell{1}{2} = {c=2}{},
  cell{1}{4} = {c=2}{},
  cell{1}{6} = {c=2}{},
  vline{4,6} = {2}{},
  vline{2,4,6} = {3-6}{},
  hline{1,3-7} = {-}{},
  hline{2} = {2-7}{},
}
Setting       & front-left       &            & front       &              & front-right       &            \\
              & \textit{Translation}      & \textit{Rotation}   & \textit{Translation} & \textit{Rotation}     & \textit{Translation}       & \textit{Rotation}   \\
\textit{Set1} & (0.5, 0.5, 0)    & (0, 0, 0)  & (1, 0, 0)   & (0, 0, 0)    & (0.5, -0.5, 0)    & (0, 0, 0)  \\
\textit{Set2} & (0.5, 0.5, -0.5) & (0, 0, 0)  & (1, 0, 0)   & (0, 0, 0)    & (0.5, -0.5, -0.5) & (0, 0, 0)  \\
\textit{Set3} & (0.5, 0.5, -0.5) & (10, 0, 0) & (1, 0, 0)   & ( 0, -10, 0) & (0.5, -0.5, -0.5) & (-10, 0,0) \\
\textit{Set4} & (0, 1, 0)        & (0, 0, 0)  & (0, 1, 0)   & (0, 0, 0)    & (0, 1, 0)         & (0, 0, 0)  
\end{tblr}
\caption{The table illustrates four different parameter settings for the free-view novel view synthesis, where \textit{Translation} refers to the translation component, \textit{Rotation} refers to the rotation component, the corresponding three numbers are the lengths of translation in meters, and the magnitudes of three Euler angles for rotation in degree.
}
\label{table:cam-layout}
\end{table*}

\begin{figure*}[htbp]
\centering
\includegraphics[width=1.8\columnwidth]{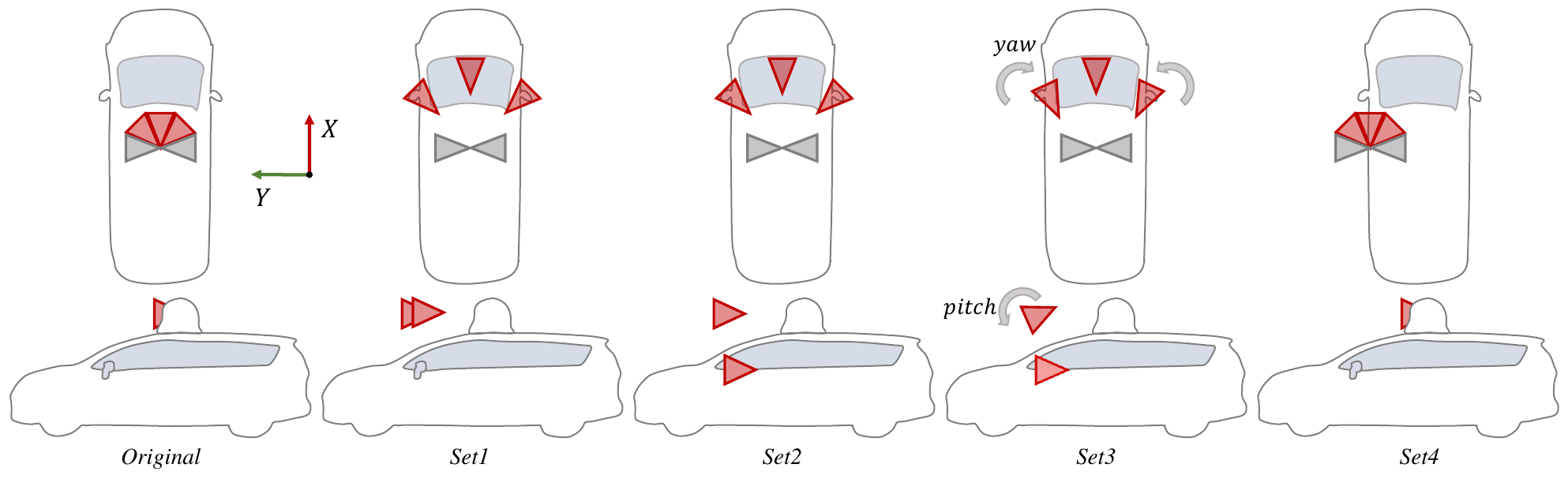} 
\caption{The camera layouts after applying the four settings of camera pose. The top row represents the camera layouts after applying various camera pose settings from a bird's eye view, while the bottom row shows the same transformations from a side view.}
\label{fig:cam-layout}
\end{figure*}

\subsection{B. Experimental results}
\paragraph{Free-view Novel View Synthesis settings.} We apply four transformations to the pose of the camera relative to the vehicle, including rotations and translations. The motive is that transferring the visual data from the collecting vehicle to a new vehicle with a diverse sensor setup is frequently demanded in AD. We set four settings based on the camera positions of the Waymo dataset to simulate the camera displacements caused by the changes in vehicle type. The detailed variation amplitudes for the front, front-left, and front-right camera under each transformation are outlined in Table \ref{table:cam-layout}, where \textit{Translation} denotes the translation distances of the camera along the $x$, $y$, and $z$ axes in meters, and \textit{Rotation} represents the rotation angles around the $yaw$, $pitch$, and $roll$ axes in degree. Figure \ref{fig:cam-layout} depicts the differences in the four transformed camera poses compared to the original Waymo configuration.

\begin{table*}[]
\centering
\begin{tblr}{
  cells = {c},
  cell{1}{1} = {r=2}{},
  cell{1}{2} = {c=2}{},
  cell{1}{4} = {c=4}{},
  vline{2,4} = {3-7}{},
  hline{1,3,5,8} = {-}{},
  hline{2} = {2-7}{},
}
Compound Mode                            & PSNR$\uparrow$ &               & FID$\downarrow$ &               &               &               \\
                                         & \textit{Train} & \textit{Test} & \textit{Set1}   & \textit{Set2} & \textit{Set3} & \textit{Set4} \\
$\text{All}_\text{3DGS}$                            & 26.91          & 25.92         & 53.12           & 61.93         & 82.52         & 61.53         \\
$\text{All}_\text{Scaffold GS}$                     & 28.04          & \textbf{26.87}         & 51.93           & 58.25         & 78.64         & 57.85         \\
$\text{E}_\text{3DGS}$ + $\text{R}_\text{3DGS}$        & 26.86          & 25.93         & 52.90           & 60.38         & 80.49         & 58.14         \\
$\text{E}_\text{Scaffold GS}$ + $\text{R}_\text{3DGS}$ & 27.91          & 26.47         & 48.95           & 58.69         & 77.92         & 57.41         \\
Ours
& \textbf{28.09}          & 26.77         & \textbf{48.05}           & \textbf{56.94}         & \textbf{75.72}         & \textbf{55.52}         
\end{tblr}
\caption{The comparison of different compound modes of Gaussians on the Waymo dataset, with each metric averaged across all clips. $\text{ALL}\_\ast$ delegates representing the scene with one united Gaussian. $\text{E}\_\ast$ and $\text{R}\_\ast$ separately means modeling the environment and road with a specific Gaussian method. As for the road part modeled by 3DGS or 2DGS of each mode, we impose the same SDF regularization. Note that all the implementations in the table are initialized from a point cloud with the sky sphere enabled. }
\label{tab:assembly}
\end{table*}

\begin{figure}[ht]
\centering
\includegraphics[width=0.99\columnwidth]{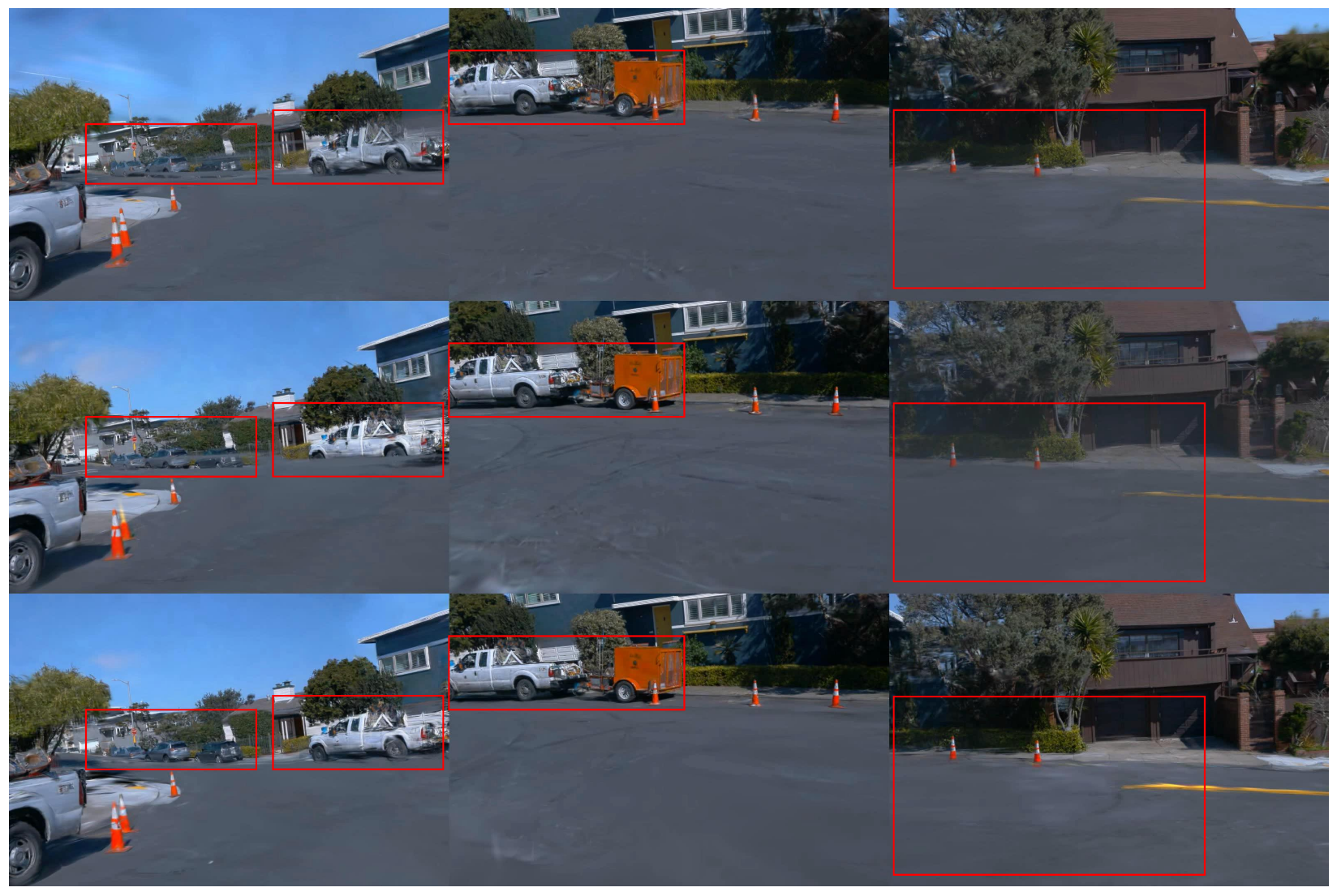} 
\caption{Qualitative comparisons of different compound modes on the Waymo dataset, with top ($\text{E}_\text{3DGS}$ + $\text{R}_\text{3DGS}$), middle ($\text{E}_\text{Scaffold GS}$ + $\text{R}_\text{3DGS}$), and bottom (Ours). The highlighted regions demonstrate that our method achieves the best rendering quality.}
\label{fig:supplement1}
\vspace{-10pt}
\end{figure}

\begin{figure*}[hbp]
\centering
\includegraphics[width=1.99\columnwidth]{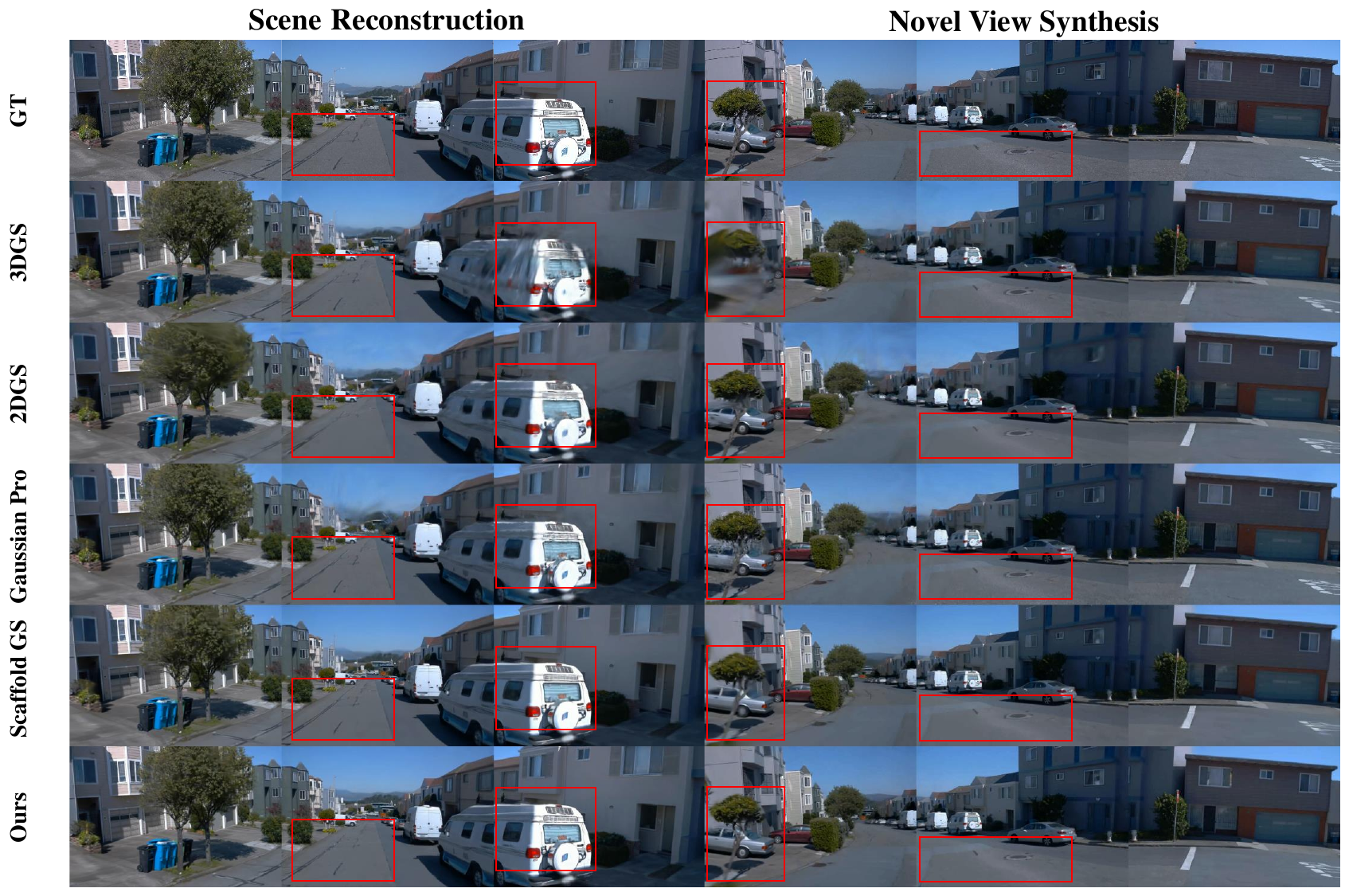} 
\caption{Qualitative comparisons on the Waymo dataset, the highlighted regions demonstrate that our method achieves the best rendering quality for both scene reconstruction in the left column and novel view synthesis in the right column.}
\label{fig:app-rebuild&nvs}
\vspace{-10pt}
\end{figure*}
\begin{figure*}[htbp]
\centering
\includegraphics[width=1.99\columnwidth]{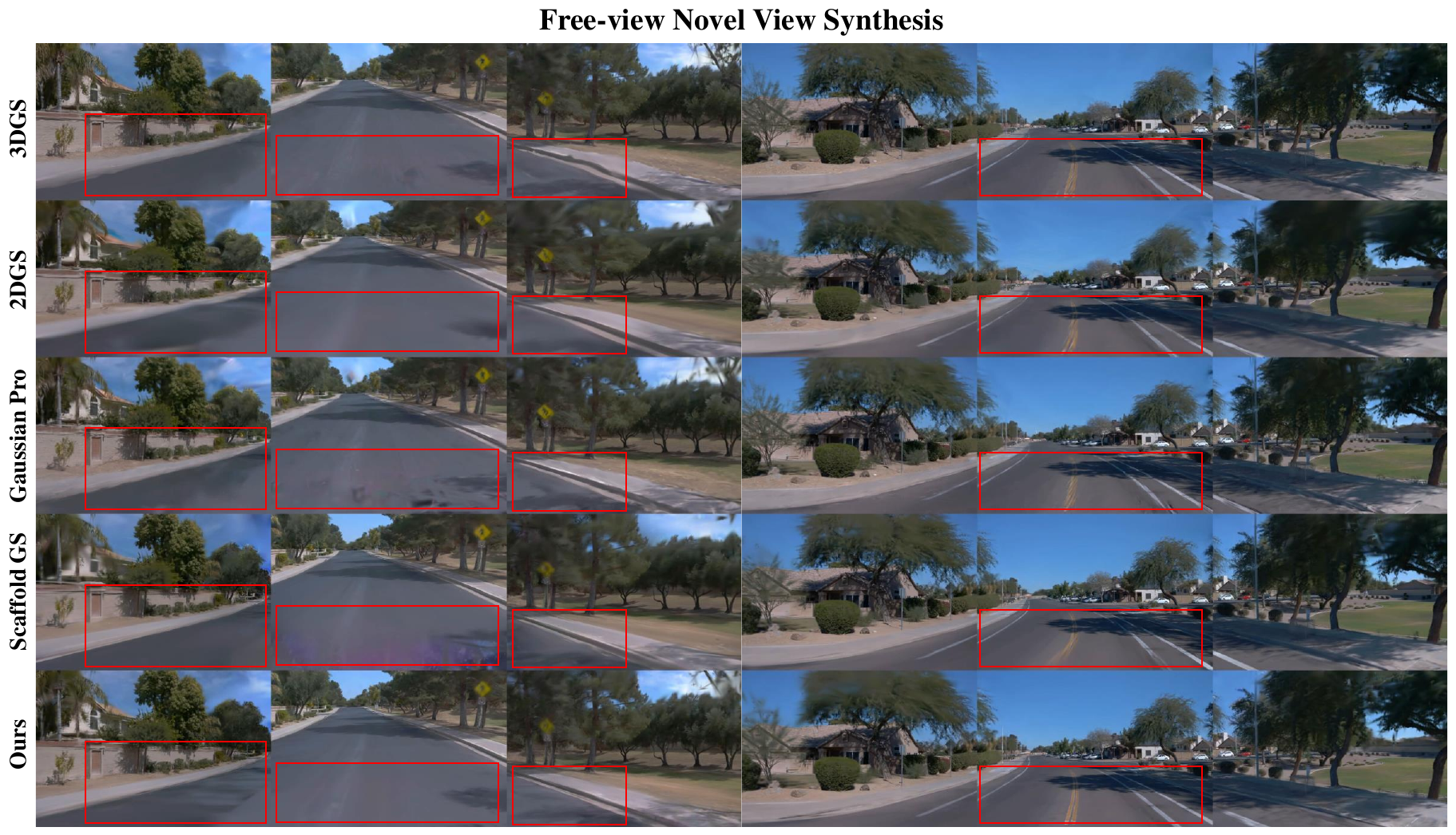} 
\caption{Qualitative results of free-view novel view synthesis under different scenarios, with the left column presenting the outputs from \textit{Set3} and the right column showing those from \textit{Set4}. The highlighted red boxes demonstrate that our method excels in capturing fine details both in terms of geometry and appearance.}
\label{fig:app-free1}
\end{figure*}
\begin{figure*}[htbp]
\centering
\includegraphics[width=1.99\columnwidth]{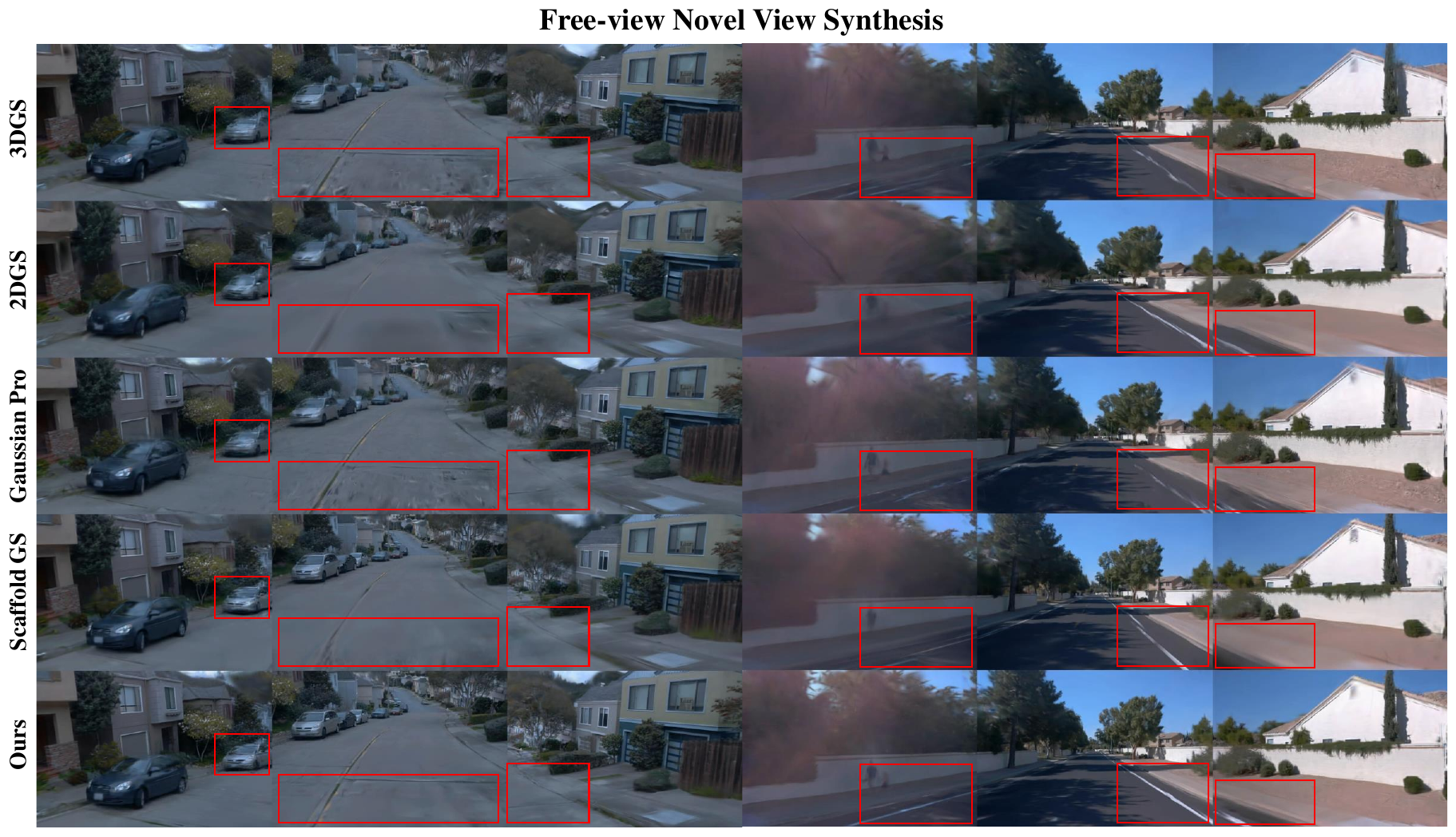} 
\caption{Qualitative results of free-view novel view synthesis under different scenarios, with the left column presenting the outputs from \textit{Set3} and the right column showing those from \textit{Set4}. The highlighted red boxes demonstrate that our method excels in capturing fine details both in terms of geometry and appearance.}
\label{fig:app-free2}
\end{figure*}

\vspace{-10pt}
\paragraph{The qualitative results on the Waymo dataset.} In addition to the quantitative metrics and qualitative rendered images elaborated in the main document, we attach additional qualitative results to provide a more intuitive performance evaluation of each method. As shown in Figures \ref{fig:app-rebuild&nvs}, \ref{fig:app-free1}, and \ref{fig:app-free2}, the qualitative results indicate that our method achieves the best visual quality in the modeling of both environmental backgrounds and road structures, demonstrating superior capability in scene reconstruction, novel view synthesis, and free-view novel view synthesis.
\vspace{-10pt}
\paragraph{Comparative analysis of the compound modes of Gaussians.} We compare several distinct assembly ways of road and non-road models, as detailed in Table \ref{tab:assembly} subsequently. Figure \ref{fig:supplement1} illustrates qualitative comparisons of three compound modes. It can be concluded that leveraging Scaffold GS and 2DGS to represent the environment and road model severally harvests the optimal performance. Benefiting from the robustness of Scaffold GS in changing views and the inherent superiority
of flat 2DGS representing the road surface, DHGS achieves superior rendering results both in the road and environment areas, exhibiting fewer floaters on the road and capturing finer environment details.


\end{document}